\documentclass[10pt]{article} 
\usepackage[accepted]{tmlr}

\usepackage{amsmath,amsfonts,bm}

\def\eqref#1{equation~\ref{#1}}

\def\1{\bm{1}}

\DeclareMathAlphabet{\mathsfit}{\encodingdefault}{\sfdefault}{m}{sl}
\SetMathAlphabet{\mathsfit}{bold}{\encodingdefault}{\sfdefault}{bx}{n}

\DeclareMathOperator*{\argmax}{arg\,max}

\usepackage{subcaption}

\usepackage{hyperref}
\usepackage{url}
\usepackage{enumitem}
\usepackage[table]{xcolor}
\usepackage{xspace,mfirstuc,tabulary}
\usepackage{graphicx,wrapfig}

\newcommand{\rllb}{\textsc{PLLB}\xspace}
\newcommand{\eg}{e.g., }

\title{Policy Learning with a Language Bottleneck}

\author{\name Megha Srivastava \email megha@cs.stanford.edu \\
      \addr 
      Stanford University
      \AND
      \name Cédric Colas \email ccolas@mit.edu \\
      \addr Massachusetts Institute of Technology \\  Inria
      \AND
      \name Dorsa Sadigh \email dorsa@cs.stanford.edu\\
      \addr Stanford University
      \AND 
      \name Jacob Andreas \email jda@mit.edu \\ \addr Massachusetts Institute of Technology}

\def\month{02}  
\def\year{2026} 
\def\openreview{\url{https://openreview.net/forum?id=sK8uEqzQPv}} 

\begin{document}

\maketitle

\begin{abstract}
Modern AI systems such as self-driving cars and game-playing agents can achieve superhuman performance, but often lack human-like generalization, interpretability, and inter-operability with human users. 
This paper introduces \textit{Policy Learning with a Language Bottleneck} (\rllb), a framework enabling AI agents to generate linguistic rules that capture the high-level strategies underlying rewarding behaviors. \rllb alternates between a \textit{rule generation} step guided by language models, and an \textit{update} step where agents learn new policies guided by rules. Crucially, \rllb enables this kind of language-guided learning even when a natural language rule is insufficient to completely describe the target policy. Across five diverse tasks, including a two-player signaling game, maze navigation, image reconstruction, and robot grasp planning, we show that \rllb learns more interpretable and generalizable behaviors than standard policy learning methods. 
In three additional human subject studies, we show that show the learned rules significantly improve \emph{human} task performance, enabling more effective human-AI coordination.\footnote{We provide source code for our experiments at \url{https://github.com/meghabyte/bottleneck}.}
\end{abstract}

\begin{figure}[h] \centering \includegraphics[width=0.9\textwidth]{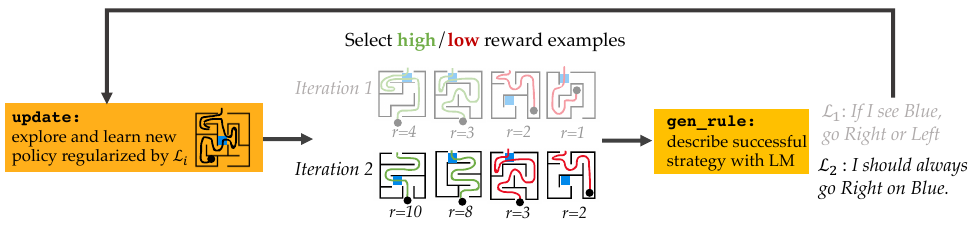} \caption{\textit{Policy Learning with a Language Bottleneck} (\rllb) alternates between two steps: 1)~\texttt{\textbf{gen\_rule}} generates a linguistic rule $\mathcal{L}_i$ explaining the agent's best behaviors by prompting a language model with contrastive (positive and negative) episodes; 2)~\texttt{\textbf{update}} learns a new policy conditioned on $\mathcal{L}_i$. \rllb strengthens human-AI coordination by constraining policies to be more interpretable and generalize better. \label{fig:overview}} \end{figure} 

\section{Introduction}

As AI systems play an increasingly central role in automation and decision-making, their success depends not only on their performance, but also on their ability to model and align with human behavior. To be truly effective assistants, these systems must act and generalize in ways that align with human expectations. However, many of today's AI systems do not meet these standards. Self-driving cars or game-playing agents like AlphaZero may achieve super-human performance, but they lack interpretability \citep{mcilroy2020chess} and often act unpredictably, especially outside their training distribution \citep{wang2023adversarial}.

Unlike AI systems that are trained in isolation, humans acquire most of their skills and knowledge from interacting with others, often using language\,---\,via instructions, advice, or explanations that improve their decision-making capabilities \citep{carruthers_language_1998, mesoudi2018cumulative}. Language acts as a \emph{communicative} medium, enabling us to teach, learn from, and coordinate with others to solve complex problems. It also supports other \emph{cognitive} functions: even when not used for communication, it allows us to represent abstract concepts \citep{hesse1988cognitive, lakoff2008metaphors}, and plan \citep{vygotsky1965thought, carruthers_magic_1998}; it guides our attention \citep{waxman1994development,yoshida_sound_2003}, and prompts relational thinking \citep{gentner2002relational}.

Consider a driver learning to navigate novel social conventions (\eg triangle-shaped stop signs). While adapting to the environment, they might verbalize strategies to themselves to avoid future mistakes (\eg\ \textit{If the sign is triangular, I should stop}, a cognitive use), or transmit this convention to others (\eg\ telling a friend \textit{In Japan, stop signs are triangles}, a communicative use). Representing learned information in language helps humans solve problems and transmit knowledge by effectively capturing abstract problem structures that facilitate learning and generalization \citep{boutonnet2015words, chopra2019first, tessler2021learning}. Importantly, language remains useful even when it cannot fully encapsulate an entire strategy (e.g., a drivers' reflexive actions, or fine-grained driving mechanics). 

There exist many recent examples in the literature of AI systems leveraging language-based representations or linguistic feedback, but these often rely on external supervision: humans in the loop or hard-coded feedback functions (\citealp{luketina2019survey, colas2022language}). We argue that more human-like AI systems should not only \textit{use} language-based supervision but also \textit{generate their own language-based feedback} to leverage both the communicative and cognitive functions of language. Importantly, this capability should also extend to tasks that require updating an underlying policy that is only partially expressible in natural language, such as low-level control or reflexive actions. 

This paper introduces \textit{Policy Learning with a Language Bottleneck} (\rllb), a framework that provides artificial embodied agents the ability to generate linguistic rules that capture the strategies underlying their most rewarding behaviors. As shown in Figure \ref{fig:overview}, \rllb alternates between a \texttt{\textbf{rule generation}} step that explains the agent's experiences by prompting a language model (LM) with contrastive episodes, and a \texttt{\textbf{policy update}} step that learns a new policy guided by these rules. Unlike past work that solely leverages LMs for modeling agent behavior and multistep reasoning \citep{park2023generative, wei2023chainofthought}, \rllb is applicable even when aspects of the target policy cannot be expressed with language. 

\textit{Policy Learning with a Language Bottleneck} can be applied to a wide range of agent types, from RL policies to LLM-based learners to robot pose estimators, and the core mechanism of \rllb remains the same: using contrastive reward signals to extract linguistic rules via an LLM, and then conditioning policy updates on these rules depending on the type of learning agent. We investigate the role of \rllb in shaping more human-like policies across five distinct tasks. \textbf{They perform better}: in two image reconstruction tasks, \rllb agents generate instructions increasing the listeners' performance compared to non-linguistic baselines (Section~\ref{sec:builder}), and \rllb agents also help more efficiently learn robot grasping policies (Section~\ref{sec:robots}). \textbf{They improve few-shot generalization}: in a maze task, \rllb rules uncover abstract problem structure that improve learning similar mazes (Section~\ref{sec:maze}), and \rllb also reduces reliance on non-generalizable visual features in robotic manipulation (Section~\ref{sec:robots}). \textbf{They are more interpretable}: in a coordination task, agents converge on humans' preferred policy when multiple optimal policies exist (Section~\ref{sec:selectsay}). \textbf{They are more inter-operable}: in maze and image reconstruction tasks, humans achieve better rewards when interacting with \rllb agents compared to agents trained without a bottleneck (Sections~\ref{sec:maze} and \ref{sec:builder}).

\section{Background \& Related Work}

\rllb is inspired by the dual use of language as both a communicative and cognitive tool for decision-making in both humans  \citep{carruthers_language_1998} and machines \citep{colas2022language}. 

\noindent \textbf{Language for communication.} Language facilitates cooperation and coordination between humans and machines via instructions \citep{hermann_grounded_2017, chevalier2018babyai}, advice \citep{watkins2021teachable}, explanations \citep{zhong_rtfm_2020, lampinen_tell_2021}, or the formation of conventions \citep{hawkins2019continual, hu2023instructrl}. Such communicative functions increase the fidelity and breadth of cultural transmission\,---\,a  process of social learning that underlies human ecological success \citep{mesoudi2018cumulative}. \rllb agents not only learn from language, but also generate their own to be shared with others. 

\noindent \textbf{Language for cognition.} Language also augments a learner's cognitive abilities. Language-augmented RL agents represent more abstract goals \citep{jiang_language_2019}, generalize better \citep{hill_emergent_2019, colas_language_2020, wong_leveraging_2021}, explore more efficiently \citep{colas_language_2020, tam2022semantic, klissarov2023motif} and can decompose complex goals into simpler ones \citep{chen_ask_2021, ahn2022can, hu2022humanai, sharma2021skill, hu2023thought}. Our work extends these benefits to agents that learn from \textit{self-generated} linguistic feedback. 

\noindent \textbf{Inner speech.} Generating linguistic rules for oneself is a form of \textit{inner speech}. In the Vygotskian tradition, inner speech is seen as the internalization of the social speech generated by caretakers to help children solve problems \citep{vygotsky1965thought, luria1959directive}. As a result, it is thought to support our capacities for complex, long-term behaviors \citep{vygotsky1965thought, luria1959directive, hermer2001language, spelke2003makes}.  AI agents endowed with forms of inner speech (explanations, descriptions or subgoals) have been found to perform and generalize better than agents trained with purely neural representations \citep{wong_leveraging_2021, lampinen_tell_2021, roy2022explainability, hu2023thought, kim2020vehicle}. Unlike these approaches, \rllb approach generates language in an unsupervised way and maximizes downstream performance as well as interpretability and inter-operability. 

\noindent \textbf{Multi-step reasoning and text agents.} Recent works have proposed guiding LMs' reasoning by prompting them to step through sequences of ``thoughts'' \citep{wei2023chainofthought, yao2022react, li2023chain, shinn2024reflexion}.  Our approach similarly uses a language bottleneck to concisely express intermediate information useful for later behavior (i.e. policy learning). However, unlike text-based reasoning agents such as ReAct \citep{yao2022react}, Rememberer \citep{zhang2023large}, and Reflexion \citep{shinn2024reflexion}, \textbf{\rllb does not require the underlying policy to be fully expressible in text}. For example, we show that a text-only maze solving agent completely fails, whereas \rllb can improve classical Q-learning style agents (Section \ref{sec:maze}).  Moreover, instead of describing actions for a particular state, \rllb rules capture high-level strategies over a sequence of states and actions, enabling generalization. We also show the learned rules can be shared to improve human task performance across multiple user studies (Sections \ref{sec:maze} and \ref{sec:builder}).

\section{The Language Bottleneck} \label{sec:setup}
\begin{figure*}[t]
    \centering
    \includegraphics[width=0.95\textwidth]{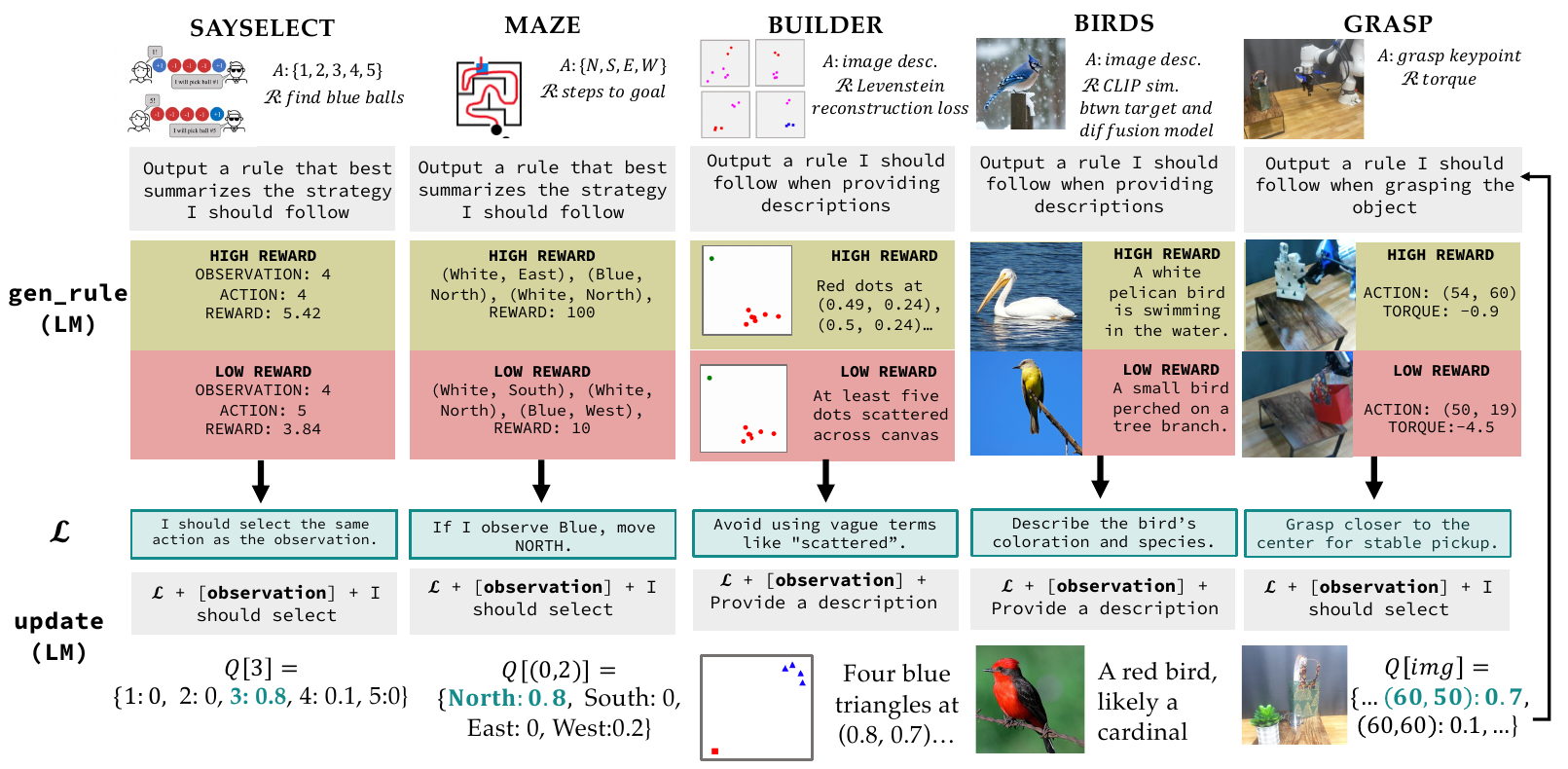}
    \caption{\rllb can be applied to a diversity of domains. In all domains, it iterates between \texttt{gen\_rule}, a function that prompts an LM (top gray boxes) to extract a linguistic rule ($\mathcal{L}$, blue) by contrasting high and low reward episodes from past experience (green and red boxes), and \texttt{update}, a function for updating an agent's policy through interaction with the environment conditioned on $\mathcal{L}$. \rllb can be applied to multi-step decision making tasks and visual or robotics tasks alike with minimal implementation variations (replacing the LM by a VLM in \texttt{gen\_rule}, or replacing policy regularization by simple instruction conditioning in  \texttt{update}. Gray boxes represent prompts \texttt{gen\_rule} and \texttt{update}, and full prompt details are in Appendix Section \ref{app:prompts}.
    }
    \label{fig:tasks}
\end{figure*}

\rllb builds on the standard RL framework to train agents to solve decision-making tasks {in human-like ways}. We formalize decision-making tasks (\eg\ solving a maze) as Markov decision processes $(S, A, f, R, T)$ with reward function $R: S \times A \rightarrow \mathbb{R}$ (\eg\ solving speed) over states $S$ (\eg\ cell coordinates) and actions $A$ (\eg\ directions), finite time horizon $T$, and a deterministic transition function $f: S \times A \rightarrow S$ that maps state--action pairs $(s, a)$ to next states $s$ (\eg\ moving between cells). Standard RL then involves training a policy $\pi$ to maximize expected reward (\eg\ learning to solve the maze faster). This is usually done by alternating between two steps: (1)~collecting data with the current policy: $D\leftarrow\pi_i$ and (2)~updating the policy using the data: $\pi_{i+1}\leftarrow\texttt{update}(\pi_i, D)$, where \texttt{update} implements an RL algorithm \citep{sutton2018reinforcement}. This procedure often fails to yield interpretable, human inter-operable, or generalizable behaviors. 

Our key idea is to introduce a \textit{language bottleneck} between data collection and policy update. We extract linguistic rules explaining past rewarding behaviors and then use them to regularize the policy's behavior in the next learning iteration (\eg \textit{I should go right in blue cells}, see Figure~\ref{fig:overview}). Our experiments will show that this improves the interpretability and inter-operability (communicative use), and performance and generalization (cognitive use) of our agents over standard RL baselines in a variety of tasks (see Figure~\ref{fig:tasks} and Sections~\ref{sec:selectsay} to \ref{sec:builder}). The resulting algorithm thus alternates between three steps: (1) data collection, $D_i\leftarrow\pi_i$, which is implemented just as in ordinary policy learning approaches; (2) language bottleneck generation, $\mathcal{L}_i \leftarrow \texttt{gen\_rule}(D_i)$ and (3) policy updating, $\pi_{i+1}\leftarrow\texttt{update}(\pi_i, D_i, \mathcal{L}_i)$ (see Figure \ref{fig:overview}). 

The core mechanism underlying \rllb is consistent across all instantiations: \texttt{gen\_rule} prompts a language model with \textit{contrastive} examples (high- vs.\ low-reward trajectories) to extract a rule $\mathcal{L}$ capturing what distinguishes successful from unsuccessful behavior, and \texttt{update} conditions the agent's subsequent learning on $\mathcal{L}$. What varies across domains is only how \texttt{update} incorporates the rule, which is depends on the learning agent's modality in that domain (e.g. Q-value regularization (Sections~\ref{sec:selectsay}--\ref{sec:maze}), prompt conditioning  for LMs (Section~\ref{sec:builder}) or visuomotor control (Section~\ref{sec:robots}). We next describe these the \texttt{gen\_rule} and \texttt{update} steps in more detail.

\subsection{Rule Generation (\texttt{gen\_rule})}
Using all the experience $D_i$ collected by the policy $\pi_i$ in the current iteration,  \texttt{gen\_rule} aims to infer an abstract rule $\mathcal{L}_i$ that best explains the agents' successful behaviors: $\mathcal{L}_i \leftarrow \texttt{gen\_rule}(D)$. This is done by prompting an LM with contrastive episodes from $D_i$ (top-$N$ highest vs. top-$N$ lowest total rewards) and asking it to \textit{provide the rule that should be followed to obtain high rewards} (see first row of Figure~\ref{fig:tasks} and Appendix Section~\ref{app:prompts} for full prompts). Importantly, this requires the first iteration of  \texttt{gen\_rule} to start only once we observe a pair episodes with sufficiently different rewards. We found this contrastive approach, inspired by \citet{zhong2023goal} and \citet{dunlap2023describing}, to provide more precise rules than simply summarizing high-reward strategies. 


\subsection{Rule-Guided Policy Update (\texttt{update})}
Given a rule $\mathcal{L}_i$, the \texttt{update} step produces a new policy $\pi_{i+1}\leftarrow\texttt{update}(\pi_i, D_i, \mathcal{L}_i)$ that is better aligned with $\mathcal{L}_i$. 
There exist many methods for leveraging language instructions to update agents' policies, though these methods traditionally focus on instructions provided by human experts. Crucially, which method to implement \rllb with depends on the underlying agent and action space representation, and will improve with advances in multi-modal modeling.  

For RL policies, we leverage InstructRL \citep{hu2023instructrl}, which regularizes the learned policy with another policy induced by the linguistic rule $\pi_\mathcal{L}$. For instance, in the maze example shown in Figure \ref{fig:overview}, if the rule is \textit{I should go right on every blue cell}, the induced policy $\pi_\mathcal{L}$ should assign probability 1 to the \textit{right} action in blue cells, but equal probabilities to all actions in every other situations. In the Q-learning algorithm, this approach simply adds a \textit{regularizing term} (orange) to the standard Q-learning update rule \footnote{This regularizing term can also be applied in the function approximation setting \citep{mnih2013playing}.}: \vspace{-2em}

{
\begin{align*}
\nonumber
    &Q^\theta(s_t, a_t) \leftarrow r_\text{t+1} + \gamma Q^\theta (s_{t+1}, a_{t+1}) ~~~ \textrm{ where}
    ~~~ a_{t+1}= \argmax_a ~ [Q(s_\text{t+1}, a) ~\textcolor{orange}{+~\lambda \log \pi_\mathcal{L}(a \mid s_t)}].
\end{align*}
}

\noindent
Here $\gamma$ is a discount factor and $\lambda$ controls the strength of the rule-induced regularization and could be made time-dependent with a pre-defined or learned schedule (\eg stronger regularization early). 

But how do we induce $\pi_\mathcal{L}$ from $\mathcal{L}$? Since $\mathcal{L}$ is expressed in natural language, we may obtain a rule-conditional policy by prompting an LM. In particular, our experiments condition the LM on both the current rule and the current state of the agent $s_t$, and instruct the LM to generate the next action to obtain a probability distribution over admissible actions (e.g directions), as in \citet{hu2023instructrl}. While these prompted LMs may perform tasks poorly on their own (e.g.\ because of their inability to perform long-range planning or process complex visual input), the regularizer may nonetheless guide Q-learning in the right direction.
Running Q-learning updates with the regularization term from experience data gives us a new Q-table $Q^\theta_{i+1}$ from which we can derive the new policy $\pi_{i+1}$ by taking actions with maximum expected value in every state $\pi_{i+1}(s_t)=\text{max}_a Q^\theta_{t+1}(s_t, a)$. 

Finally, there exist some domains where the policy can be directly implemented by an LM or VLM, such as text-based games explored by  methods like ReAct \citep{yao2022react} or Rememberer \citep{zhang2023large}. In these cases, \texttt{update} can be implemented by conditioning the policy on the rule $\mathcal{L}$ and adding it to the prompt $\pi_\mathcal{L}$, ultimately steering the agent's: $\pi_{i+1}\leftarrow\pi_{\mathcal{L}_i}$. 
Given a new policy $\pi_{i+1}$, we can now close the loop and generate a new rule $\mathcal{L}_{i+1}$.The choice between InstructRL-style regularization and prompt conditioning depends on the action space and policy architecture. In settings with large, continuous action spaces (e.g. reasoning over free-text), conditioning text-generation on rules is likely more tractable, whereas Q-learning style updates may be more appropriate for interactive environments with feedback (e.g. game playing).  We implement \rllb with tabular RL agents in Sections \ref{sec:selectsay} and \ref{sec:maze}, with LM/VLM policies in  Section \ref{sec:builder}, and visuomotor robot controllers in  
Section \ref{sec:robots}.

\section{Experiment Set-Up}
\label{sec:experiments}

We run experiments on a diverse set of five tasks, with different agent architectures and action spaces, to showcase how \rllb can train more human-like agents. These include a simple two-player communication game called \textsc{SaySelect} (Section~\ref{sec:selectsay}), a \textsc{Maze} solving task (Section~\ref{sec:maze}), two collaborative image reconstruction tasks with synthetic (\textsc{Builder}) and natural (\textsc{Birds}) images (Section \ref{sec:builder}), and grasp planning for a 7-DoF robot (Section \ref{sec:robots}). For all tasks, we include hyperparameter details (Appendix \ref{app:hyperparams}),  the exact prompts used in $\texttt{gen\_rule}$ and $\texttt{update}$ (Appendix~\ref{app:prompts}), and examples of generated rules. For our main experiments, we prioritized using accessible and low-cost open weight language models such as \textsf{llama-2-70b-chat}. However, we analyze the effect of changing the  underlying language model  on $\texttt{gen\_rule}$  in Appendix \ref{app:rules} and \ref{app:rules-qualitative}.

To evaluate the \textit{cognitive} function of \rllb, we first evaluate the policies learned by artificial agents in each environment based on performance metrics such as task reward, similarity to known human-like policies, or generalization. For all tasks, we compare our \textbf{Bottleneck} approach with baseline methods (\eg standard RL, base LM) and an \textbf{Adversarial} ablation of \rllb that leverages invalid linguistic rules (\eg obtained by labeling high reward examples as low reward) in \texttt{gen\_rule}.  The purpose of  \textbf{Adversarial} is to test whether \rllb gains arise from meaningful rule content rather than merely introducing any linguistic signal.

To evaluate the \textit{communicative} function of \rllb, we conduct several human subject studies showing that generated rules $\mathcal{L}$ can help improve human task performance on a variety of metrics (\eg reward, solving time). Participants were either recruited in-person (for \textsc{Maze}) or on  \href{http://prolific.com}{Prolific} (\textsc{Builder}, \textsc{Birds}), compensated at a rate of US\$12/hour. All studies were approved by our institution's IRB. To the best of our knowledge, our work is the first to validate self-generated agent rules from LMs for training human participants.

Finally, we conduct additional sensitivity analysis on the affects of variations in prompt, model temperature, and non-contrastive episode selection in the Appendix.

\section{SaySelect}\label{sec:selectsay}
\begin{figure}[t]
    \centering
    \includegraphics[width=0.9\textwidth]{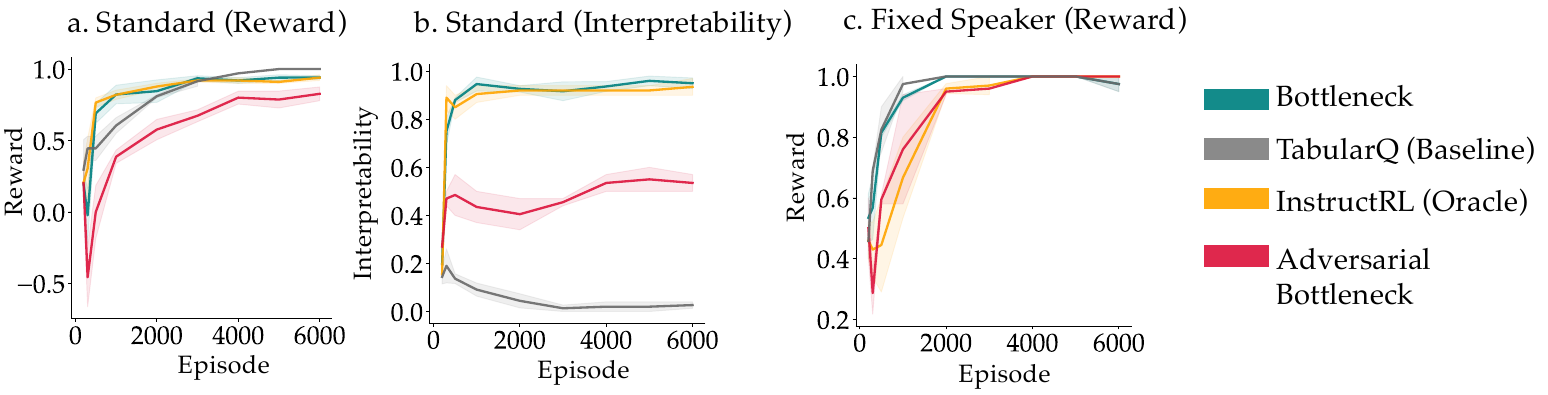}
    \caption{\textsc{SaySelect}. a) \textbf{Bottleneck} agents learn as fast as \textbf{TabularQ} and \textbf{InstructRL} and faster than agents using adversarial rules \textbf{Adversarial}. b) Unlike \textbf{TabularQ}, they learn human-interpretable policies; this without relying on an external instruction like \textbf{InstructRL}. c) When faced with speakers enforcing a non-human-interpretable policy, \textbf{Bottleneck} converges faster than \textbf{InstructRL}.}
    \label{fig:ss_results}
\end{figure}

As a proof-of-concept, we first consider \textsc{SaySelect}, a simple collaborative game introduced in \citet{hu2023instructrl}. Here, a \textit{speaker} can see a hidden set of five balls including two blue ones and must help a \textit{listener} find the two blue balls by communicating with them only via numbers, see Appendix Figure~\ref{fig:ss-overview}. In principle, the speaker and listener could converge on any bijective convention associating messages to balls, and this is what we find with RL agents (multiple optimal policies). Humans, on the other hand, empirically prefer the mapping 1 $\rightarrow$ 1, 2 $\rightarrow$ 2, etc.\, which we call the \textit{human-interpretable policy}.  Whereas \citet{hu2023instructrl} showed that regularizing the listener's policy with the instruction \textit{I should select the same number as my partner} successfully guided the two RL agents to the human-interpretable policy, we ask whether \rllb could learn a similar rule and achieve a similar interpretability \textit{on its own}?

\noindent \textbf{Task Overview.}
The speaker and listener are RL agents trained with Q-Learning that receive positive rewards when the listener collects blue balls, negative rewards otherwise. After training both agents for 200 episodes, we prompt \textsf{llama-2-70b-chat}\footnote{https://huggingface.co/meta-llama/Llama-2-70b-chat-hf} with the task description and a contrastive set of high- and low-reward episodes to generate a rule $\mathcal{L}$, \eg \textit{I should choose action 4 whenever the observation is 2, 3, 4, or 5.} (\texttt{gen\_rule}, see Figure~\ref{fig:tasks}). 

We generate new rules every 500 training episodes. In between, we regularize the listener's policy with a another policy $\pi_\mathcal{L}$ induced from the current rule $\mathcal{L}$. This is done by applying a \textit{softmax} on the LM's logits (see method in Section~\ref{sec:setup}) across all possible actions and obtaining a distribution. In the following experiments, we compare our \textbf{Bottleneck} method, with the \textbf{Adversarial} version (corrupted rule), a \textbf{TabularQ} baseline, as well as an \textbf{InstructRL} upper bound using the same predefined instruction as \citep{hu2023instructrl}.

\subsection{\rllb helps learn human-interpretable policies}
We first compare the four methods along two dimensions: their reward and their \textit{interpretability} measured as the proportional (0 to 1) similarity between the listener's policy and the optimal human-interpretable policy (the intuitive message-ball described above). All methods but \textbf{Adversarial} quickly solve the task (Figure~\ref{fig:ss_results}a), showing that corrupted rules hinder learning. \textbf{Bottleneck} and \textbf{InstructRL} both converge on the human-interpretable policy thanks to their linguistic rules while \textbf{TabularQ}\,---\,deprived of such rule\,---\,does not (Figure~\ref{fig:ss_results}b). Interestingly,  \textbf{Adversarial} policies are more interpretable than \textbf{TabularQ} ones, suggesting that even corrupted language may provide inductive biases towards more human-like behavior. Qualitatively, we observe that generated rules converge towards describing the human-interpretable policy (\eg\ \emph{I should follow the strategy of choosing the same action as Agent 1}, see examples in Section \ref{app:rules}).

\subsection{\rllb can learn counter-intuitive policies}
A potential confound in the preceding experiment is the possibility that \rllb converges to a human-interpretable policy because this is the \emph{only} behavior an LM can describe. To test this hypothesis, we fixed the speaker to use a counter-intuitive policy using a random mapping between balls and messages (\eg\ 1 $\rightarrow$ 3, 5 $\rightarrow$ 2, etc.). Here, the human-interpretable policy is mis-aligned, and would lead the listener to fail or learn slower. In Figure \ref{fig:ss_results}, we observe that while \textbf{InstructRL} and \textbf{Adversarial}, both regularized by misaligned rules (the human-interpretable one for \textbf{InstructRL}, another random permutation for \textbf{Adversarial}), eventually adapt to the fixed speaker, they converge at a slower rate than both \textbf{TabularQ} (no rule) and \textbf{Bottleneck}. This effect was more pronounced at higher values of $\lambda$. Overall, \textbf{Bottleneck} is the only method that adapts to a fixed speaker policy while also converging on the human-like policy when the speaker is learning, without human supervision.

\section{Maze}\label{sec:maze}
\begin{figure}[t]
    \centering
    \includegraphics[width=0.9\textwidth]{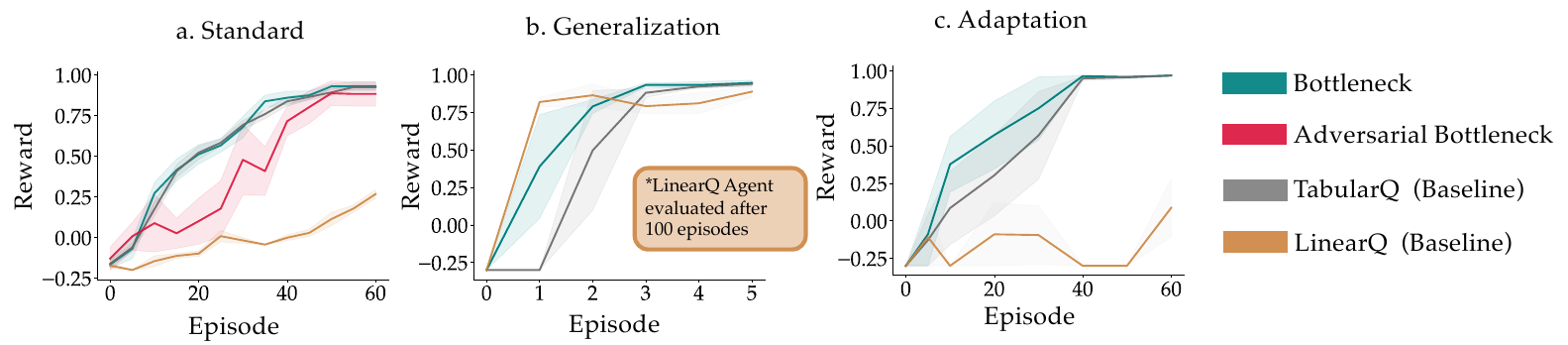}
    \caption{Results in \textsc{Maze}. a) \textbf{Bottleneck} agents learn as fast as the non-linguistic \textbf{Baseline agents}, but faster than \textbf{Adversarial} and \textbf{LinearQ} agents. b) When faced with a new maze with similar structure, \textbf{Bottleneck} agents learn faster than \textbf{TabularQ} (which does not perceive color) and \textbf{LinearQ} (which does, but learns slower). c) When faced with a maze of a different structure, \textbf{Bottleneck} agents adapt swiftly while \textbf{LinearQ} cannot recover. We do not evaluate \textbf{Adversarial} for the Generalization or Adaptation experiments due to its poor performance in the standard setting.} 
    \label{fig:maze-results}
\end{figure}

In the \textsc{Maze} domain, agents must navigate a maze to find the goal using four directional actions (N/S/E/W). We study mazes in which environment cues provide hints about the optimal path. This setup helps us study the impact of the language bottleneck: can \rllb agents infer the underlying structure and use it to generalize across mazes (cognitive use)? Can they transmit this knowledge to humans and help them perform better in novel settings (communicative use)?

\noindent \textbf{Task Overview.} We randomly generate different 7x7 mazes with the \textsf{gym-maze} code base.\footnote{https://github.com/MattChanTK/gym-maze} Agents receive as reward the inverse of the number of steps they needed to reach the goal as an incentive to learn the optimal solution. Although all mazes are random, we introduction additional structures by coloring cells for which the optimal action is \textit{south} (red) or \textit{north then east} (blue) with probability 50\% (or they are left blank). Therefore, agents that leverage color information can better generalize to new mazes where the coloring is preserved, even if the optimal action sequence differs. 

The first iteration of \rllb corresponds to training a standard RL algorithm to obtain $\pi_1$, in our case a tabular Q-learning agent (\textbf{TabularQ}). After the agent observes at two solved mazes, we run \texttt{gen\_rule} by prompting a \textsf{llama-2-70b-chat} LM with contrastive episodes and a task description. This gives us a linguistic rule $\mathcal{L}_1$ (\eg \textit{Upon observing RED, take SOUTH}) that we can use to \texttt{update} the policy. We first induce the regularizing policy $\pi_{\mathcal{L}_1}$ with rule $\mathcal{L}_1$ by obtaining a probability distribution over the 4 actions from the LM, and then run the RL algorithm for 5 episodes to obtain the new policy $\pi_{2}$ (procedure described in Section~\ref{sec:setup}). We repeat these steps every 5 episodes of interactions with the environment. 

Baselines include the tabular Q-learning algorithm without language bottleneck (\textbf{TabularQ}) and a variant of \textbf{Bottleneck} generating rules from reward-randomized episode samples (\textbf{Adversarial}). We also evaluate \textbf{LinearQ}, an agent learning a linear model Q-function with an additional feature for cell color. We train \textbf{LinearQ} with a batch size of 10 and learning rate 0.001, after performing a hyperparameter sweep. We found \textbf{LinearQ} took significantly longer to converge than the other methods, likely due to the increased state representational complexity.

\subsection{\rllb improves few-shot generalization} We next evaluate \rllb's ability to improve the few-shot generalization capabilities for policies on unseen mazes with a similar underlying structure. For a a fair comparison, we start with the fully converged policy for each method (requiring 100 episodes for \textbf{LinearQ}). 
Figure~\ref{fig:maze-results}a shows that, in the Standard setting, learning a valid rule (\textbf{Bottleneck}) does not increase learning speed of a single maze over \textbf{TabularQ}. However, using a corrupted rule (\textbf{Adversarial}) or learning from linear features (\textbf{LinearQ}) slows down learning. Furthermore, \textbf{Bottleneck} outperforms \textbf{TabularQ} with respect to few-shot generalization: when we  switch to a new 7x7 maze sharing the same underlying color semantics, but with a different optimal action sequences. \textbf{Bottleneck} leverages the learned rule and generalizes to the new maze more effectively than all other agents (Figure~\ref{fig:maze-results}b).
While \textbf{TabularQ} cannot generalize because it does not perceive colors, \textbf{LinearQ} does generalize faster at first, but converges slower. 

The generated rules improve over time to better capture the underlying structure of the maze (\eg \textit{if I observe BLUE, then take the NORTH action}; see other examples in Appendix ~\ref{app:rules}). Across all trials, $100\%$ of the final rules mention the red $\rightarrow$ \textit{south} rule and $60\%$ uncovered the more complex blue $\rightarrow$ \textit{north-then-east} rule. Finally, we find that a policy that does not update Q-values (that is, purely implemented via the LM) achieves a success rate $0\%$ within the same time limit, even when using the final rules generated from \rllb. This highlights the importance of leveraging experience that cannot be expressed with language, which text-based reasoning agents cannot do.

\subsection{\rllb learns adaptable policies}

In Figure~\ref{fig:maze-results}c, we reproduce this same experiment on a maze with a different underlying structure (red now indicates \textit{west} while blue indicates \textit{east then south}). Although the rule \textbf{Bottleneck} learned does not apply anymore, it can still adapt faster than the baselines.  We find that all trials end with rules capturing the new mapping (100\% red $\rightarrow$ \textit{west}, 50\% blue $\rightarrow$ \textit{east-then-south}). Meanwhile, \textbf{LinearQ} overfits to the first maze and struggle to adapt.  Overall, our results show that \rllb supports human-like cognitive functions by finding a tradeoff between the efficiency of TabularQ and the generalization capabilities LinearQ, while remaining more adaptable.

\subsection{\rllb is more interpretable and inter-operable}
\begin{wrapfigure}{R}{0.4\textwidth}
\vspace{-27px}
\centering\includegraphics[width=0.4\textwidth]{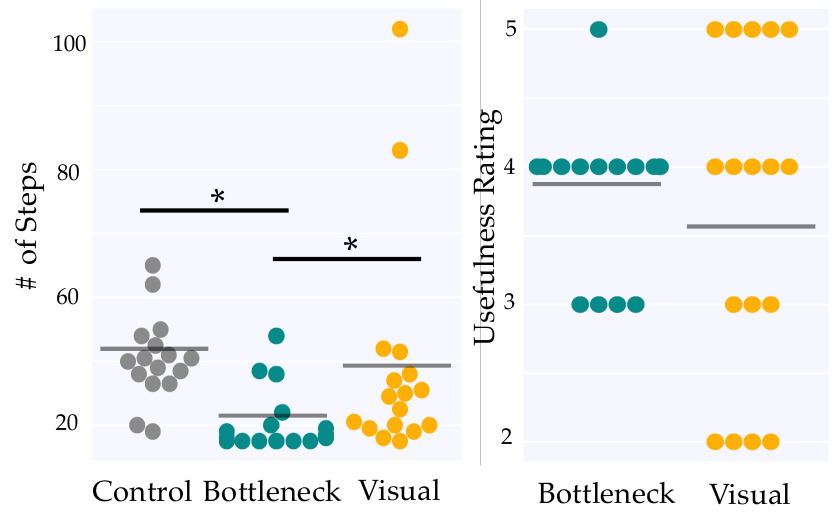}
    \caption{Participants given \rllb rules solve new mazes faster than those given either visual or no aid (control). They also self-reported linguistic feedback as more useful than  visual. * marks statistical significance at level 0.05 with two-sided Mann-Whitney test after Bonferroni correction. }
    \label{fig:maze-human}
\end{wrapfigure}
Can generated rules be useful to humans as well? We asked 50 university students to solve a 7x7 maze in the fewest steps possible using arrow keys. They could only observe the cells they had already visited and the walls they had already bumped into (see Figure \ref{fig:maze-interface} in Appendix). We split them into three groups: a \textbf{Control} group receiving no assistance, and two others receiving information about a \textit{similar} but different maze sharing the same underlying color semantics. The \textbf{Visual} group is shown a visual representation of the optimal policy (arrows in each cell, see Figure~\ref{fig:maze-visualaid} in Appendix), while the \textbf{Bottleneck} group is provided a randomly sampled language rule $\mathcal{L}$ learned by \rllb. This set-up lets us evaluate how generalizable the two different aids are to a new maze, as well as how quickly participants can account for incorrect information depending on modality. 

Participants using \rllb rules solve the new maze with fewer steps than others and find this aid significantly more useful than the visual one in average (see Figure~\ref{fig:maze-human}). Participants indeed found it harder to extract the relevant information from the visual aid, which contained a great deal of non-transferable information (optimal actions in all non-colored cells). \rllb rules focus on the useful and transferable information learned by the previous agent, which is much more readily usable by humans. Overall, the results in \textsc{Maze} demonstrate the ability of \rllb to train agents that are more generalizable and adaptable (cognitive use) as well as more interpretable by, and inter-operable with humans (communicative use).

\section{Collaborative Image Reconstruction}\label{sec:builder}
\begin{figure}[h]
    \centering
    \includegraphics[width=0.9\textwidth]{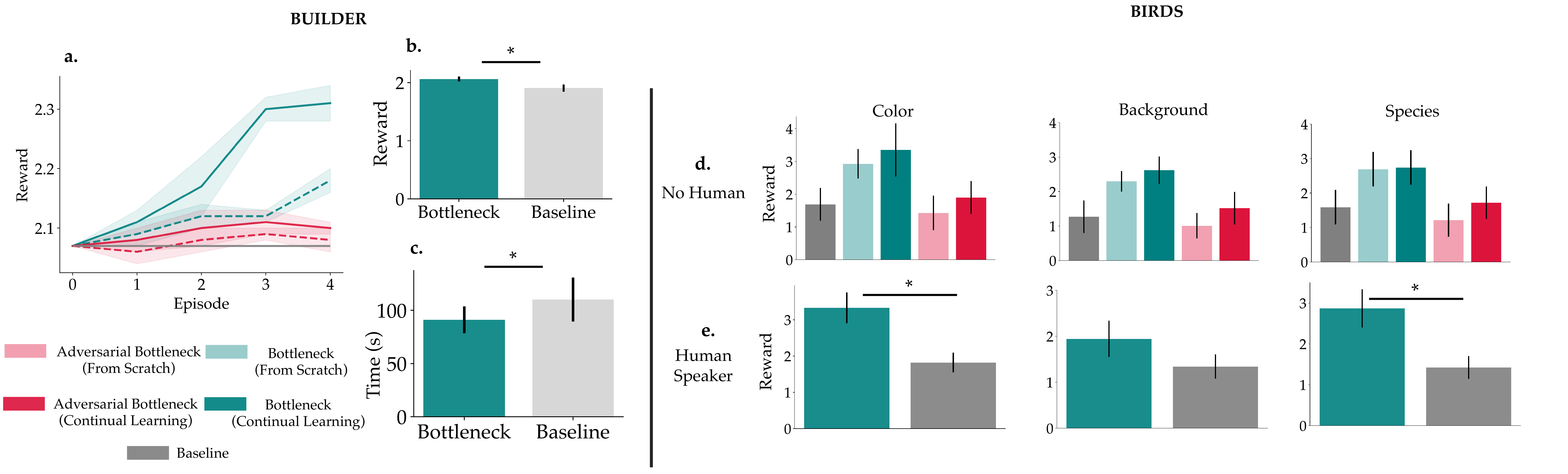}
    \caption{\textsc{Builder}:  \textbf{Bottleneck} listeners improve upon \textbf{Baseline} descriptions over time, unlike an (\textbf{Adversarial}) listener that inaccurately represents rewards. Human users reconstructed target images with  b. higher accuracy, and  c. less time,  when provided \textbf{Bottleneck} instruction vs. original \textbf{Baseline} descriptions. \textsc{Birds}: \rllb  rules (3 iterations) help d. automated and e. human speakers generate better descriptions for naturalistic image reconstruction for three distinct reward functions.}
    \label{fig:builder-automatic}
\end{figure}

\rllb is not restricted to the training of RL agents. This section introduces two collaborative image reconstruction tasks inspired by the collaborative assembly task of \citet{mccarthy202assembly}. \textsc{Builder} and \textsc{Birds} both consider two agents: a \textit{speaker}, who can see a hidden target image, and a \textit{listener}, who must accurately reconstruct the target image based on a description from the speaker. Both agents aim to converge to a description style leading to high reconstruction accuracy. We consider both synthetic images built from sequences of discrete actions by the listener (\textsc{Builder}), and natural images of birds that the listener reconstructs using a text-to-image generation model (\textsc{Birds}). Here, \rllb combines with state-of-the-art multi-modal models to let agents perceive semantic visual features. 

\subsection{Task Overview}
\paragraph{Dataset.} In \textsc{Builder}, we construct a dataset of synthetic images with a variable number of shapes (triangle, square, circle) in different colors (magenta, blue, red, green) and locations on a 2D grid. Each image is created by a sequence of discrete actions, each representing a particular combination of shape, color, and location (\eg \texttt{[ACT12 ACT2 ACT4]}, where ACT2 adds a magenta circle in the lower right area of the grid). In \textsc{Birds}, we construct a dataset of natural bird images by selecting 5 images for each of 10 bird species from the CUB-200-2011 dataset provided by \citet{cub2011birds} (see Appendix Section \ref{app:birds_dataset} for species list). 

\paragraph{Reward.} In \textsc{Builder}, since target and reconstructed images are fully determined by their action sequences, we measure task success via the Levenshtein similarity between the (sorted) corresponding sequences. In contrast, the \textsc{Birds} listener outputs text-to-image generations with varied properties, so we define three reward functions: \textbf{Color}, \textbf{Background}, and \textbf{Species}. To evaluate a reconstructed image $x_r$ against a target $x_t$, we form a \textit{contrast} set $C$ from  three CUB-200-2011 images differing in the target property (\eg color) but most similar in others. Reward is then $r = \sum_{c\in C}d(\textsf{CLIP}(x_r),\textsf{CLIP}(x_c)) - d(\textsf{CLIP}(x_r),\textsf{CLIP}(x_t))$, where \textsf{CLIP} denotes the CLIP image embedding \citep{radford2021learning}. For both \textsc{Builder} and \textsc{Birds}, rewards are reported on a held-out validation set.

\paragraph{Models.} As open-source VLMs struggle with describing synthetic images, we use the \textsf{
llama-2-70b-chat} LM as our speaker for \textsc{Builder}, representing images in raw text that list all shapes' type, color, and exact coordinates individually in sequence. On the other hand, we use the open source \textsf{Llava}
VLM as our speaker for \textsc{Birds} \citep{liu2023improved}. We prompt both speaker models using the target image and a general task description, as well generated rules after the first iteration.

We implement the listener for \textsc{Builder} with a neural sequence-to-sequence model pre-trained on English text \citep{lewis2019bart}, which we fine-tune at each episode on a training set of (description, action sequence) pairs using  descriptions provided by the speaker. Meanwhile, we use  the \textsf{Stable Diffusion} text-to-image diffusion model as the listener for \textsc{Birds}.

For both tasks, we consider two settings: \textbf{From Scratch}, where the listener is initialized using the original pre-trained model weights at every episode, and \textbf{Continual Training}, where the listener is continuously updated over time using descriptions by the speaker. We create separate held-out splits for training, early stopping, selecting samples for \texttt{gen\_rule}, and evaluation.

\paragraph{Implementation.} We first generate base image descriptions for both tasks by providing the speaker a prompt containing a general task description and target image, but without any rule $\mathcal{L}$.
For  the \textsc{Builder} task, an original image description is: \textit{At least five dots in total - four red and one green - scattered across the canvas, with two sets of matching locations (x = 0.49 / y = 0.24 and x = 0.5 / y = 0.24), another set at x = 0.52 / y = 0.3, and finally, one dot located near x = 0.76 / y = 0.71.} This description is overly complex,  mixes different coordinate formats, and difficult for the listener to use to reconstruct the target image, leading to low reward. Meanwhile, an example original image description for the \textsc{Bird} task is: 
\textit{``a bird on a branch''}. While simple, this description does not contain sufficient information about the \textbf{Color}, \textbf{Background}, and \textbf{Species}, causing the listener to generate an image reconstruction with low reward. 

We next evaluate these base descriptions using each task's respective reward functions, and select the (image, description) examples with the 5 highest and lowest rewards. Using the prompts shown in Section \ref{app:prompts}, we 
implement \texttt{gen\_rule} using the \textsf{llama-2-70b-chat} and \textsf{llava} models for \textsc{Builder} and \textsc{Birds}, respectively.
For \textsc{Birds}, we observe that output rules $\mathcal{L}$ are specific to the reward function: an example rule we get for the \textbf{Species} reward is: \textit{``Identify  the  bird's  species  if  possible,  and  include  any  distinctive  characteristics  that  set  it  apart  from  other  birds.''}, while an example rule we get for the \textbf{Background} reward is: \textit{``Describe  the  setting  or  background  of  the  image,  such  as  the  presence  of  snow,  water,  or  other  elements''}. 

We implement \texttt{update} for both tasks by simply appending the output rule $\mathcal{L}$ to the original prompt to the speaker, as shown in Figure \ref{fig:tasks}, using the speaker's output as the new description for a given target image. We repeat this cycle of eliciting rule $\mathcal{L}$ ,  appending $\mathcal{L}$ to the same prompt used to generate the base instructions in order to re-label our data with new instructions, and training and evaluating the listener model for a total of 5 iterations for \textsc{Builder} and 3 iterations for \textsc{Birds}. We qualitatively show this leads to improved image reconstruction from the listener in Appendix Figure \ref{fig:builder-bird-overivew},  and next discuss our experimental results.

\subsection{\rllb helps speakers provide more usable instructions}
 For both tasks, listeners following rules generated by \rllb (\textbf{Bottleneck}) outperform listeners trained from uninformed (\textbf{Baseline}) or misinformed (\textbf{Adversarial}) speakers (Figure \ref{fig:builder-automatic}). This holds true for all three different reward functions in  \textsc{Birds}, as well as both training settings, with \textbf{Continual Learning} learning enabling \textbf{Bottleneck} to have an even stronger improvement over \textbf{Baseline} image descriptions by leveraging learned task experience.  \textbf{Continual Learning} does not significantly help the \textbf{Adversarial}  method, emphasizing that the linguistic rules $\mathcal{L}$ in \textbf{Bottleneck} do capture succesfull task strategies. However,  \textbf{Continual Learning} is not required for \textbf{Bottleneck} to outperform \textbf{Baseline} image descriptions. The success of the \textbf{From Scratch} training setting can be interpreted as the speaker generating abstract rules to guide its own learning, leading to improved performance and showing evidence for a cognitive use of language.

Finally, we observe that the generated rules $\mathcal{L}$ encourage the speaker to reduce vague and general language, and describe properties relevant to the corresponding underlying reward function in \textsc{Birds}, although this might take several cycles to fully converge (see example rules in Appendix Section~\ref{app:rules}). Furthermore, rules $\mathcal{L}$ become more complex over time, and across different trials we observe path dependency where different trials converge to different rules (\eg $(x,y)$ vs. $x=0.xx, y=0.yy$ for coordinates in \textsc{Builder}). Overall, these rules help guide the speaker and listener agents in both tasks to iteratively improve image reconstruction over time, which can also be seen qualitatively in Appendix Figure~\ref{fig:builder-bird-overivew}.

\subsection{\rllb can collaborate with human listeners}
Do human listeners benefit from image descriptions provided by speakers following \textbf{Bottleneck} rules? We  conduct a human subject study for the \textsc{Builder} task,  where we replace the listener with study participants at the end of one iteration of \rllb, and evaluate how quickly they can reconstruct target images from a held-out test set. We recruit 20 crowdworkers on Prolific to reconstruct 5 images using descriptions from a speaker  following a rule $\mathcal{L}$  generated with (\textbf{Bottleneck}),  and 5 images using the original (\textbf{Baseline}) descriptions. We provided participants an interface that included a drawing canvas and buttons controlling shapes and colors (Section \ref{app:figs}), allowing us to use 
 Levenshtein similarity between (sorted) user actions and the underlying action sequences for a target image as reward.

Participants using \textbf{Bottleneck} descriptions built target images faster and more accurately than participants using control descriptions  (Wilcoxon signed-rank test, $p<0.05$, Figure \ref{fig:builder-automatic}). In a post-study survey, 55\% preferred \textbf{Bottleneck} descriptions, describing them as \textit{``more direct and less ambiguous''}. 10\% found both descriptions types equally easy to follow while remaining participants preferred \textbf{Baseline} descriptions because  \textit{they gave more flexibility}, leading to a more unstructured user study. However, we note that that user preference for customization does not correlate with any notion of improved  performance in our original goal of image reconstruction.

\subsection{\rllb collaborates with human speakers} 
Can humans benefit from rules generated by \rllb? We asked 12 Prolific crowdworkers to act as speakers and provide descriptions for our dataset of \textsc{Birds} images. Half of the participants were provided rules corresponding to one of the three reward functions ($\mathcal{L}_3$) while the other half were not. Figure~\ref{fig:builder-automatic} (bottom) shows that listeners instructed by human speakers provided with \rllb rules outperform those instructed by uninformed human speakers (\textbf{Baseline}) (Bonferroni-corrected t-tests $p<0.05$ for \textbf{species} and \textbf{colors}). 

The descriptions from users not provided a rule are more diverse and less focused \eg \textit{A barbed wire is an uncomfortable stop for a bird}. With \rllb rules on the other hand (\eg \textit{Describe the bird’s coloration accurately}), they generate more specific descriptions: \textit{Red and black bird on barbed wire. Bright red chest, red at top of head, black wings and beak} for the same target image (see other example rules in Appendix~Table~\ref{tab:diffusion_chains}). \rllb agents can easily transmit what they learned from experience (the rule) to humans.

\section{Robot Grasp Planning}
\label{sec:robots}
\begin{figure*}[t]
    \centering
    \includegraphics[width=0.7\textwidth]{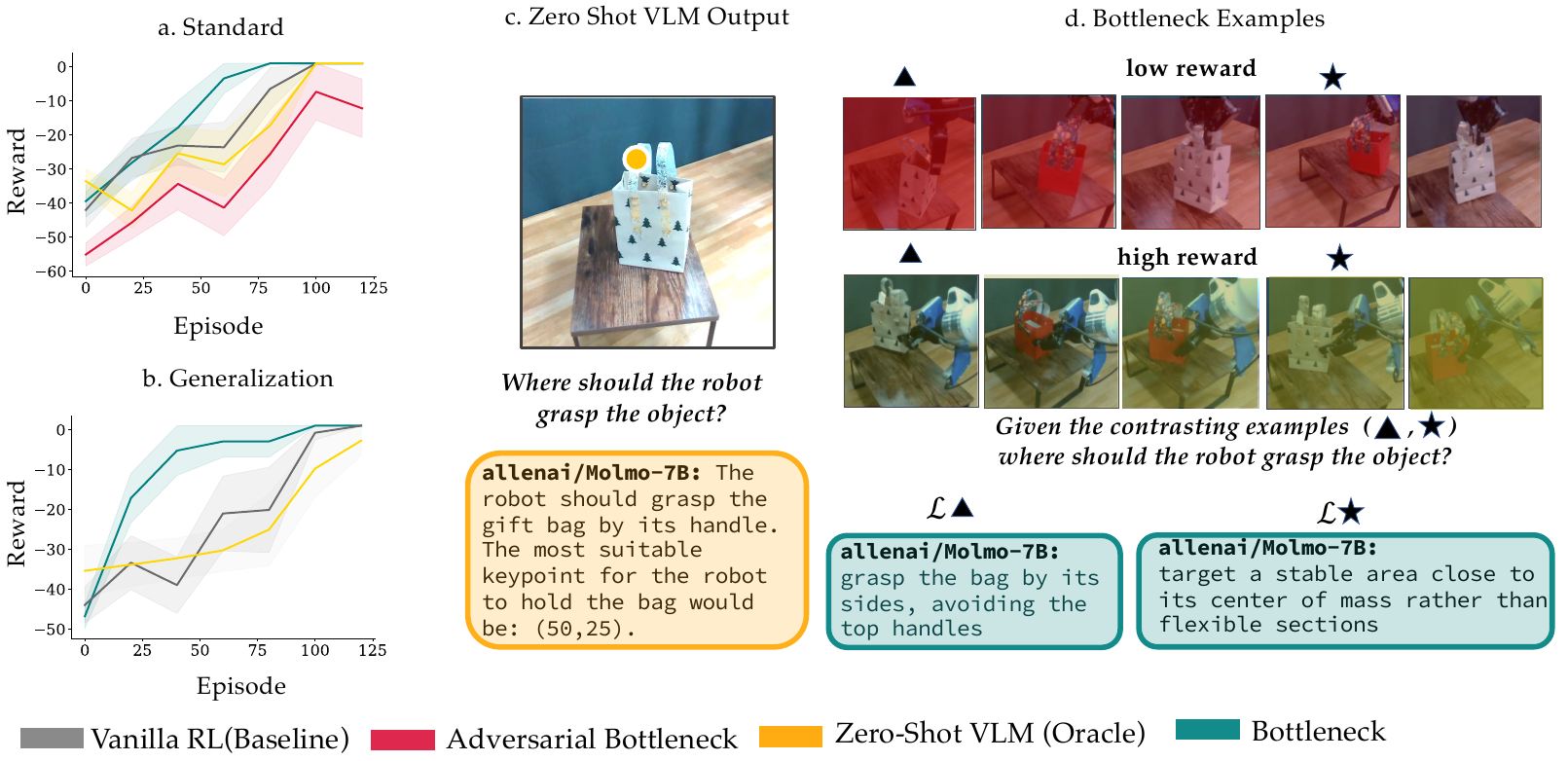}
    \caption{Results in \textsc{Grasp}. a) \textbf{Bottleneck} agents learn faster than the non-linguistic RL \textbf{Baseline agents}, as well as \textbf{Adversarial} and an \textbf{Oracle} zero-shot prompt of a VLM without any internal policy update. b) \textbf{Bottleneck} agents generalize faster than both \textbf{Baseline} (which over-relies on image details) and \textbf{Oracle} to unseen bag designs and locations, (c) shows the fixed linguistic rule d) Example grasps, shaded by reward. } 
    \label{fig:robot-results}
\end{figure*}

Can \rllb rules improve policy learning in  embodied tasks with complex action spaces such as robot planning? Existing works on LM-based agents (e.g., ReAct \citep{yao2022react} or Reflexion \citep{shinn2024reflexion}) and robot policies (e.g., Code as Policies) leverage pre-trained models for action generation, but cannot be applied in domains where a (non-LM-based) policy must be updated using real-world RL signals.


Consider training a robot to lift a paper bag. The most intuitive grasp would be at the bag's handles; indeed, a zero-shot \textbf{Oracle} prompt of the \textsf{molmo-7b} multi-modal language model outputs grasp keypoints on bag handles (Figure \ref{fig:robot-results}).  However, if the bag is too heavy, a robot should adapt to less common grasps (e.g. by the side) to avoid tearing. Because VLMs are not grounded with actual physical quantities, such adaptation is impossible without leveraging an external reward \citep{gao2024physically}. On the other hand, existing RL-based approaches for robotic manipulation are often inefficient \citep{kalashnikov2018qt}. We show that \rllb rules can be used to efficiently learn non-intuitive robot grasping policies. 

\subsection{Task Overview} In the \textsc{Grasp} task, our agent is a Franka Emika Panda Arm robot that needs to lift and move a paper bag filled with unknown objects off of a shelf, as shown in Appendix Figure \ref{fig:robot-setup}. Due to the complex action space, we follow \citet{kalashnikov2018qt} and decompose the robot's policy into (i) perceiving the scene and identifying an appropriate grasp location, and (ii) planning a path to this location. 

We begin with a \emph{fixed} keypoint-conditioned path-planning policy which is pre-trained on a dataset of 50 trajectories with 3D-point cloud observations using  \textsf{Sphinx} \citep{sundaresan2025whats}. This model outputs a trajectory of 7-DoF end-effector poses (x/y/z/yaw/pitch/row/gripper) for the robot to execute given the keypoint.
We then use \rllb to train (via RL) a grasp prediction policy that selects a grasp keypoint (a 2-D position) given an image of the environment.
We define a successful grasp as one where the robot can lift the bag without it breaking, and so set our reward to be 0 for a successful grasp, -100 if the bag is not grasped at all, and use the  normalized downward torque (e.g. -20Nm) applied to the robot's shoulder joint as a dense reward for unsuccessful grasps. 
Initially, the agent performs random grasps.
After it observes one successful grasp, we run $\texttt{gen\_rule}$ with the open-source \textsc{molmo-7b} vision--language model with contrastive images of successful and unsuccessful grasp keypoints. This gives us a linguistic rule $\mathcal{L}$ (e.g.\ \textit{grasp the left side of the bag}) that we can use to $\texttt{update}$ the policy. We regularize the initial grasping policy toward the policy $\pi_\mathcal{L}$ induced from rule $\mathcal{L}$ by prompting \textsf{molmo-7b} while conditioning on $\mathcal{L}$, then instructing it to generate random keypoints. By executing \textsf{Sphinx}  with these keypoints, the overall policy is updated to condition on $\mathcal{L}$.

We compare our \textbf{Bottleneck} method with an \textbf{Adversarial} version (swapping high and low reward grasps and an \textbf{Oracle} that always guides the policy with the rule generated from zero-shot prompting \textsf{molmo-7b} (no constrasting examples based on a reward).  We call this Zero-Shot VLM an \textbf{Oracle} as it represents the VLM's privileged prior knowledge about intuitive grasping strategies, or knowledge a human expert would also possess.  We also evaluate a \textbf{Vanilla RL} baseline where the policy only uses image observation data to predict a grasp keypoint based on task reward after execution, which helps us isolate PLLB's incremental value for keypoint selection. 

\subsection{\rllb can learn counter-intuitive policies} As shown in Figure \ref{fig:robot-results}, when averaged across 20 random seeds, our \textbf{Bottleneck} converges to higher reward compared to all other methods. Because the \textbf{Oracle} method suggests a rule encouraging the policy to predict key points near the bag's handles, it requires more episodes to learn that this behavior leads to lower reward. Qualitatively, we observe the generated rules converge towards guiding the robot to grasp at less common locations (e.g., \textit{Grasp the bottom of the object}, see examples in Appendix Section \ref{app:rules}). This shows that \rllb can help improve over standard RL methods even for learning counter-intuitive policies. 

\subsection{\rllb improves few-shot generalization}
To evaluate  few-shot generalization, we create an environment with a new bag, location, and set of visual distractions (see Appendix \ref{fig:robot-setup}). We then evaluate \textbf{Bottleneck}, \textbf{Zero-Shot VLM (Oracle)},  and \textbf{Vanilla RL (Baseline)} when intialized with the policies learned in the original environment, allowing for updates from the environment over time. As shown in Figure \ref{fig:robot-results}, the rules learned by \textbf{Bottleneck} outperform alternative methods for generalizing to this new environment. Because the \textbf{Vanilla RL} policy can only rely on image features, it is not able to leverage the more generalizable strategy \rllb rules capture and select grasp keypoints that hold semantic meaning (e.g. the bottom of the bag) captured by pre-trained LMs. Overall, the \textsc{Grasp} results show that \rllb can improve policy convergence even for the complex dynamics of embodied tasks.

\section{Limitations \& Discussion}\label{sec:limitations}

\rllb helps agents becomes more interpretable by and inter-operable with humans. 
In our current implementation, \texttt{rule\_gen}  requires converting episodes into LM-compatible representations (text/images), which may limit use with long-horizon sensorimotor trajectories. Nevertheless, our Grasp task successfully uses pretrained robotic models, and advances in long-context modeling can address this limitation. 

One natural question is whether \rllb's gains arise specifically from language, or from reward-aligned abstraction more generally? Comparing against non-linguistic bottlenecks (e.g., learned latent codes) is challenging: such representations require training a model across many task variations to become meaningful, whereas language provides pre-structured abstractions via pretrained LLMs that can immediately be used with single tasks. Furthermore, our \textbf{Adversarial} ablations across experiments confirm that rule content, not merely the presence of a bottleneck, drives improvements with \rllb. 

Our human studies also demonstrate uniquely linguistic benefits (interpretability, human transfer) that would be unavailable with other latent representations (including code).
However, rule generation can  fail when the LM either abstracts too much (\eg in \textsc{Birds} we once found the rule: \textit{``Use  descriptive  language  that  conveys  the  mood,  such  as  the  serenity  of  a  snowy  day  or  the  freedom  of  flight''}) or, more seriously, causes harm in safety-critical situations by creating a false trust in generated rules.  We believe improvements in multi-modal models can help mitigate these issues (Appendix \ref{app:rules-qualitative}), as well as additional human-in-the-loop verification steps of rules for high-stakes situations.  

Our work opens up a set of interesting questions around designing intelligent sampling of contrastive episodes in \texttt{rule\_gen}. For example, sampling episodes that are not outliers can  help with handling stochastic environments, where high reward trajectories may not necessarily be optimal. Sampling of contrastive episodes can also adjust to known properties of how LMs attend to long context (i.e. more attention to episodes at the start and end).  Another interesting direction for future work is applying \rllb to tasks with complex reward functions that reflect hard-to-articulate human preferences, such as image captioning for accessibility \citep{nie2020pragmatic} and personalizing language models \citep{li2023eliciting}. Although one concern is whether the linguistic rules from \rllb might exacerbate spurious reward signals, any reward misspecification becomes more detectable and diagnosable through interpretable rules, whereas black-box policies offer no such transparency.

Our experiments with \rllb focus on environments with tractable  action spaces (e.g. discrete choices, text captions, keypoints for robotic grasping). In open-ended text generation like math reasoning or code synthesis, this approach would require further adaptation. Possible extensions include: (1) using the rule as a verifier or reward bonus rather than an action prior, (2) discretizing the action space into semantically meaningful categories that the LM can reason over, or (3) using the rule to filter or re-rank candidate actions from a base policy. Such adaptations would extend the cognitive and communicative benefits of language-guided learning to a broader class of sequential decision-making problems \citep{colas2022language}, and we leave exploration of these extensions to future work.

\section{Acknowledgements}
We would like to acknowledge support from NSF Awards \#2006388, \#2125511, \#2238240 and \#2212310, AFOSR YIP, ONR \#N000142112298, and DARPA. 
Research was sponsored by the Department of the Air Force Artificial Intelligence Accelerator and was accomplished under Cooperative Agreement Number FA8750-19-2-1000. The views and conclusions contained in this document are those of the authors and should not be interpreted as representing the official policies, either expressed or implied, of the Department of the Air Force or the U.S. Government. The U.S. Government is authorized to reproduce and distribute reprints for Government purposes notwithstanding any copyright notation herein.
Megha Srivastava was additionally supported by an IBM PhD Fellowship. Cédric Colas received funding from the European Union’s Horizon 2020 research and innovation program under the
Marie Skłodowska-Curie grant agreement No.\ 101065949.

\bibliography{main}

@article{zhang2023large,
  title={Large Language Models Are Semi-Parametric Reinforcement Learning Agents},
  author={Zhang, Danyang and Chen, Lu and Zhang, Situo and Xu, Hongshen and Zhao, Zihan and Yu, Kai},
  journal={arXiv preprint arXiv:2306.07929},
  year={2023}
}

@INPROCEEDINGS{kim2020vehicle,
  author={Kim, Jinkyu and Moon, Suhong and Rohrbach, Anna and Darrell, Trevor and Canny, John},
  booktitle={2020 IEEE/CVF Conference on Computer Vision and Pattern Recognition (CVPR)}, 
  title={Advisable Learning for Self-Driving Vehicles by Internalizing Observation-to-Action Rules}, 
  year={2020},
  volume={},
  number={},
  pages={9658-9667},
  doi={10.1109/CVPR42600.2020.00968}}

@misc{allenai2025molmo,
  author    = {{Allen Institute for AI}},
  title     = {MOLMO Blog},
  howpublished = {\url{https://molmo.allenai.org/blog}},
  note      = {Accessed: 2025-02-25}
}

@inproceedings{sundaresan2025whats,
  author    = {Priya Sundaresan and Hengyuan Hu and Quan Vuong and Jeannette Bohg and Dorsa Sadigh},
  title     = {What's the Move? Hybrid Imitation Learning via Salient Points},
  booktitle = {Proceedings of the International Conference on Learning Representations (ICLR)},
  year      = {2025},
  month     = {February}
}

@article{kalashnikov2018qt,
  title={QT-Opt: Scalable Deep Reinforcement Learning for Vision-Based Robotic Manipulation},
  author={Kalashnikov, Dmitry and Irpan, Alex and Pastor, Peter and Ibarz, Julian and Herzog, Alexander and Jang, Eric and Quillen, Deirdre and Holly, Ethan and Kalakrishnan, Mrinal and Vanhoucke, Vincent and Levine, Sergey},
  journal={arXiv preprint arXiv:1806.10293},
  year={2018},
  url={https://arxiv.org/abs/1806.10293}
}

@inproceedings{gao2024physically,
  title={Physically Grounded Vision-Language Models for Robotic Manipulation},
  author={Gao, Jensen and Sarkar, Bidipta and Xia, Fei and Xiao, Ted and Wu, Jiajun and Ichter, Brian and Majumdar, Anirudha and Sadigh, Dorsa},
  booktitle={Proceedings of the IEEE International Conference on Robotics and Automation (ICRA)},
  year={2024},
  url={https://arxiv.org/abs/2309.02561}
}

@misc{park2023generative,
      title={Generative Agents: Interactive Simulacra of Human Behavior}, 
      author={Joon Sung Park and Joseph C. O'Brien and Carrie J. Cai and Meredith Ringel Morris and Percy Liang and Michael S. Bernstein},
      year={2023},
      eprint={2304.03442},
      archivePrefix={arXiv},
      primaryClass={cs.HC}
}

@misc{dunlap2023describing,
      title={Describing Differences in Image Sets with Natural Language}, 
      author={Lisa Dunlap and Yuhui Zhang and Xiaohan Wang and Ruiqi Zhong and Trevor Darrell and Jacob Steinhardt and Joseph E. Gonzalez and Serena Yeung-Levy},
      year={2023},
      eprint={2312.02974},
      archivePrefix={arXiv},
      primaryClass={cs.CV}
}

@misc{mnih2013playing,
      title={Playing Atari with Deep Reinforcement Learning}, 
      author={Volodymyr Mnih and Koray Kavukcuoglu and David Silver and Alex Graves and Ioannis Antonoglou and Daan Wierstra and Martin Riedmiller},
      year={2013},
      eprint={1312.5602},
      archivePrefix={arXiv},
      primaryClass={cs.LG}
}

@misc{zhong2023goal,
      title={Goal Driven Discovery of Distributional Differences via Language Descriptions}, 
      author={Ruiqi Zhong and Peter Zhang and Steve Li and Jinwoo Ahn and Dan Klein and Jacob Steinhardt},
      year={2023},
      eprint={2302.14233},
      archivePrefix={arXiv},
      primaryClass={cs.CL}
}

@article{luketina2019survey,
  title={A survey of reinforcement learning informed by natural language},
  author={Luketina, Jelena and Nardelli, Nantas and Farquhar, Gregory and Foerster, Jakob and Andreas, Jacob and Grefenstette, Edward and Whiteson, Shimon and Rockt{\"a}schel, Tim},
  eprint={1906.03926},
  archivePrefix={arXiv},
  primaryClass={cs.CL},
  year={2019}
}

@misc{wei2023chainofthought,
      title={Chain-of-Thought Prompting Elicits Reasoning in Large Language Models}, 
      author={Jason Wei and Xuezhi Wang and Dale Schuurmans and Maarten Bosma and Brian Ichter and Fei Xia and Ed Chi and Quoc Le and Denny Zhou},
      year={2023},
      eprint={2201.11903},
      archivePrefix={arXiv},
      primaryClass={cs.CL}
}

@inproceedings{mccarthy202assembly,
  title={Emergence of compositional abstractions in human collaborative assembly},
  author={William McCarthy},
  year={2020},
  url={https://api.semanticscholar.org/CorpusID:233179096}
}

@misc{tessler2021learning,
      title={Learning to solve complex tasks by growing knowledge culturally across generations}, 
      author={Michael Henry Tessler and Jason Madeano and Pedro A. Tsividis and Brin Harper and Noah D. Goodman and Joshua B. Tenenbaum},
      year={2021},
      eprint={2107.13377},
      archivePrefix={arXiv},
      primaryClass={cs.CL}
}

@article{chopra2019first, 
author = {Chopra, Sahil and Tessler, Michael Henry and Goodman, Noah D}, 
journal = {Proc. of CogSci}, 
title = {The first crank of the cultural ratchet: Learning and transmitting concepts through language}, 
year = {2019} 
}

@article {boutonnet2015words,
	author = {Bastien Boutonnet and Gary Lupyan},
	title = {Words Jump-Start Vision: A Label Advantage in Object Recognition},
	volume = {35},
	number = {25},
	pages = {9329--9335},
	year = {2015},
	doi = {10.1523/JNEUROSCI.5111-14.2015},
	publisher = {Society for Neuroscience},
	URL = {https://www.jneurosci.org/content/35/25/9329},
	eprint = {https://www.jneurosci.org/content/35/25/9329.full.pdf},
	journal = {Journal of Neuroscience}
}

@book{vygotsky1965thought,
  title={Thought and Language},
  author={Vygotsky, L.S},
  year={1965},
  publisher={MIT Press}
}

@misc{wang2023adversarial,
      title={Adversarial Policies Beat Superhuman Go AIs}, 
      author={Tony T. Wang and Adam Gleave and Tom Tseng and Kellin Pelrine and Nora Belrose and Joseph Miller and Michael D. Dennis and Yawen Duan and Viktor Pogrebniak and Sergey Levine and Stuart Russell},
      year={2023},
      eprint={2211.00241},
      archivePrefix={arXiv},
      primaryClass={cs.LG}
}

@misc{liu2023improved,
      title={Improved Baselines with Visual Instruction Tuning}, 
      author={Haotian Liu and Chunyuan Li and Yuheng Li and Yong Jae Lee},
      year={2023},
      eprint={2310.03744},
      archivePrefix={arXiv},
      primaryClass={cs.CV}
}

@misc{radford2021learning,
      title={Learning Transferable Visual Models From Natural Language Supervision}, 
      author={Alec Radford and Jong Wook Kim and Chris Hallacy and Aditya Ramesh and Gabriel Goh and Sandhini Agarwal and Girish Sastry and Amanda Askell and Pamela Mishkin and Jack Clark and Gretchen Krueger and Ilya Sutskever},
      year={2021},
      eprint={2103.00020},
      archivePrefix={arXiv},
      primaryClass={cs.CV}
}

@techreport{cub2011birds,
	Title = "The Caltech-UCSD Birds-200-2011 Dataset" ,
	Author = {Wah, C. and Branson, S. and Welinder, P. and Perona, P. and Belongie, S.},
	Year = {2011}}

@misc{nie2020pragmatic,
      title={Pragmatic Issue-Sensitive Image Captioning}, 
      author={Allen Nie and Reuben Cohn-Gordon and Christopher Potts},
      year={2020},
      eprint={2004.14451},
      archivePrefix={arXiv},
      primaryClass={cs.CL}
}

@misc{lewis2019bart,
      title={{BART}: Denoising Sequence-to-Sequence Pre-training for Natural Language Generation, Translation, and Comprehension}, 
      author={Mike Lewis and Yinhan Liu and Naman Goyal and Marjan Ghazvininejad and Abdelrahman Mohamed and Omer Levy and Ves Stoyanov and Luke Zettlemoyer},
      year={2019},
      eprint={1910.13461},
      archivePrefix={arXiv},
      primaryClass={cs.CL}
}

@inproceedings{mcilroy2020chess, series={KDD ’20},
   title={Aligning Superhuman AI with Human Behavior: Chess as a Model System},
   url={http://dx.doi.org/10.1145/3394486.3403219},
   DOI={10.1145/3394486.3403219},
   booktitle={Proceedings of the 26th ACM SIGKDD International Conference on Knowledge Discovery Data Mining},
   publisher={ACM},
   author={McIlroy-Young, Reid and Sen, Siddhartha and Kleinberg, Jon and Anderson, Ashton},
   year={2020},
   month=aug, collection={KDD ’20} }

@misc{hu2022humanai,
      title={Human-{AI} Coordination via Human-Regularized Search and Learning}, 
      author={Hengyuan Hu and David J Wu and Adam Lerer and Jakob Foerster and Noam Brown},
      year={2022},
      eprint={2210.05125},
      archivePrefix={arXiv},
      primaryClass={cs.AI}
}

@misc{hu2023instructrl,
      title={Language Instructed Reinforcement Learning for Human-{AI} Coordination}, 
      author={Hengyuan Hu and Dorsa Sadigh},
      year={2023},
      eprint={2304.07297},
      archivePrefix={arXiv},
      primaryClass={cs.AI}
}

@inproceedings{yoshida_sound_2003,
 author = {Yoshida, Hanako and Smith, Linda B},
 booktitle = {Proc. of CogSci},
 title = {Sound symbolism and early word learning in two languages},
 year = {2003}
}

@book{gentner2002relational,
 author = {Dedre Gentner and Jeffrey Loewenstein},
 booktitle = {Language, Literacy, and Cognitive Development: The Development and Consequences of Symbolic Communication},
 language = {English},
 publisher = {Erlbaum},
 title = {Relational Language and Relational Thought},
 year = {2002}
}

@book{carruthers_language_1998,
 author = {Carruthers, Peter and Boucher, Jill},
 journal = {Interdisciplinary themes. Cambridge, UK},
  publisher = {Cambridge University Press},
 title = {Language and {Thought}},
 year = {1998}
}

@incollection{carruthers_magic_1998,
 author = {Clark, Andy},
 booktitle = {Language and {Thought}},
 edition = {1},
 editor = {Carruthers, Peter and Boucher, Jill},
 isbn = {978-0-521-63108-2 978-0-521-63758-9 978-0-511-59790-9},
 publisher = {Cambridge University Press},
 title = {Magic {Words}: {How} {Language} {Augments} {Human} {Computation}},
 year = {1998}
}

@book{lakoff2008metaphors,
 author = {Lakoff, George and Johnson, Mark},
 publisher = {University of Chicago press},
 title = {Metaphors We Live By},
 year = {2008}
}

@article{hesse1988cognitive,
  title={{The Cognitive Claims of Metaphor}},
  author={Hesse, Mary},
  journal={{The Journal of Speculative Philosophy}},
  year={1988},
  publisher={JSTOR}
}

@article{waxman1994development,
  title={The development of an appreciation of specific linkages between linguistic and conceptual organization},
  author={Waxman, Sandra R},
  journal={Lingua},
  volume={92},
  pages={229--257},
  year={1994},
  publisher={Elsevier}
}

@article{hill_emergent_2019,
 author = {Hill, Felix and Lampinen, Andrew K. and Schneider, Rosalia and Clark, Stephen and Botvinick, Matthew and McClelland, James L. and Santoro, Adam},
 journal = {Proc. of ICLR},
 title = {Emergent systematic generalization in a situated agent},
 year = {2020}
}

@article{mesoudi2018cumulative,
  title={What is cumulative cultural evolution?},
  author={Mesoudi, Alex and Thornton, Alex},
  journal={Proceedings of the Royal Society B},
  volume={285},
  number={1880},
  pages={20180712},
  year={2018},
  publisher={The Royal Society}
}

@article{colas_language_2020,
 author = {C{\'{e}}dric Colas and
Tristan Karch and
Nicolas Lair and
Jean{-}Michel Dussoux and
Cl{\'{e}}ment Moulin{-}Frier and
Peter F. Dominey and
Pierre{-}Yves Oudeyer},
journal = {Proc. of NeurIPS},
 title = {{Language as a cognitive tool to imagine goals in curiosity driven exploration}},
 year = {2020}
}

@article{li2023eliciting,
  title={Eliciting human preferences with language models},
  author={Li, Belinda Z and Tamkin, Alex and Goodman, Noah and Andreas, Jacob},
eprint={2310.11589},
      archivePrefix={arXiv},
      primaryClass={cs.CS},
  year={2023}
}

@article{colas2022language,
  title={Language and culture internalization for human-like autotelic AI},
  author={Colas, C{\'e}dric and Karch, Tristan and Moulin-Frier, Cl{\'e}ment and Oudeyer, Pierre-Yves},
  journal={Nature Machine Intelligence},
  volume={4},
  number={12},
  pages={1068--1076},
  year={2022},
  publisher={Nature Publishing Group UK London}
}

@article{hermer2001language,
  title={Language, space, and the development of cognitive flexibility in humans: The case of two spatial memory tasks},
  author={Hermer-Vazquez, Linda and Moffet, Anne and Munkholm, Paul},
  journal={Cognition},
  volume={79},
  number={3},
  pages={263--299},
  year={2001},
  publisher={Elsevier}
}

@article{wong_leveraging_2021,
 author = {Wong, Catherine and Ellis, Kevin and Tenenbaum, Joshua B. and Andreas, Jacob},
 journal = {Proc. of ICML},
 title = {Leveraging language to learn program abstractions and search heuristics},
 year = {2021}
}

@article{spelke2003makes,
  title={What makes us smart? Core knowledge and natural language},
  author={Spelke, Elizabeth S},
  journal={Language in mind: Advances in the study of language and thought},
  pages={277--311},
  year={2003}
}

@article{chen_ask_2021,
 author = {Chen, Valerie and Gupta, Abhinav and Marino, Kenneth},
 journal = {Proc. of ICLR},
 title = {Ask your human: Using human instructions to improve generalization in reinforcement learning},
 year = {2021}
}

@article{sharma2021skill,
 journal = {Proc. of ACL},
 author = {Pratyusha Sharma and Antonio Torralba and Jacob Andreas},
 title = {{Skill induction and planning with latent language}},
 year = {2021}
}

@article{tam2022semantic,
  title={Semantic exploration from language abstractions and pretrained representations},
  author={Tam, Allison and Rabinowitz, Neil and Lampinen, Andrew and Roy, Nicholas A and Chan, Stephanie and Strouse, DJ and Wang, Jane and Banino, Andrea and Hill, Felix},
  journal={Proc. of NeurIPS},
  year={2022}
}

@article{hermann_grounded_2017,
 author = {Hermann, Karl Moritz and Hill, Felix and Green, Simon and Wang, Fumin and Faulkner, Ryan and Soyer, Hubert and Szepesvari, David and Czarnecki, Wojciech Marian and Jaderberg, Max and Teplyashin, Denis and Wainwright, Marcus and Apps, Chris and Hassabis, Demis and Blunsom, Phil},
      eprint={1706.06551},
      archivePrefix={arXiv},
      primaryClass={cs.CV},
 title = {Grounded language learning in a simulated 3D world},
 year = {2017}
}

@article{chevalier2018babyai,
  title={BabyAI: A platform to study the sample efficiency of grounded language learning},
  author={Chevalier-Boisvert, Maxime and Bahdanau, Dzmitry and Lahlou, Salem and Willems, Lucas and Saharia, Chitwan and Nguyen, Thien Huu and Bengio, Yoshua},
  journal={Proc. of ICLR},
  year={2019}
}

@inproceedings{zhong_rtfm_2020,
 author = {Victor Zhong and
Tim Rockt{\"{a}}schel and
Edward Grefenstette},
 booktitle = {Proc. of ICLR},
 title = {{RTFM}: Generalising to New Environment Dynamics via Reading},
 year = {2020}
}

@book{sutton2018reinforcement,
  title={Reinforcement Learning: An Introduction},
  author={Sutton, Richard S and Barto, Andrew G},
  year={2018},
  publisher={MIT press}
}

@article{li2023chain,
  title={Chain of code: Reasoning with a language model-augmented code emulator},
  author={Li, Chengshu and Liang, Jacky and Zeng, Andy and Chen, Xinyun and Hausman, Karol and Sadigh, Dorsa and Levine, Sergey and Fei-Fei, Li and Xia, Fei and Ichter, Brian},
eprint={2312.04474},
      archivePrefix={arXiv},
      primaryClass={cs.CS},
  year={2023}
}

@article{yao2022react,
  title={React: Synergizing reasoning and acting in language models},
  author={Yao, Shunyu and Zhao, Jeffrey and Yu, Dian and Du, Nan and Shafran, Izhak and Narasimhan, Karthik and Cao, Yuan},
    eprint={2210.03629},
      archivePrefix={arXiv},
      primaryClass={cs.CS},
  year={2022}
}

@article{shinn2024reflexion,
  title={Reflexion: Language agents with verbal reinforcement learning},
  author={Shinn, Noah and Cassano, Federico and Gopinath, Ashwin and Narasimhan, Karthik and Yao, Shunyu},
  journal={Proc. of NeurIPS},
  year={2024}
}

@article{hawkins2019continual,
  title={Continual adaptation for efficient machine communication},
  author={Hawkins, Robert D and Kwon, Minae and Sadigh, Dorsa and Goodman, Noah D},
  journal={Proc. of ACL},
  year={2020}
}

@article{lampinen_tell_2021,
 author = {Lampinen, Andrew K. and Roy, Nicholas A. and Dasgupta, Ishita and Chan, Stephanie C. Y. and Tam, Allison C. and McClelland, James L. and Yan, Chen and Santoro, Adam and Rabinowitz, Neil C. and Wang, Jane X. and Hill, Felix},
 journal = {Proc. of ICML},
 title = {Tell me why! -- Explanations support learning of relational and causal structure},
 year = {2022}
}

@inproceedings{watkins2021teachable,
 author = {Olivia Watkins and Abhishek Gupta and Trevor Darrell and Pieter Abbeel and Jacob Andreas},
 booktitle = {Proc. of NeurIPS},
 editor = {A. Beygelzimer and Y. Dauphin and P. Liang and J. Wortman Vaughan},
 title = {Teachable Reinforcement Learning via Advice Distillation},
 year = {2021}
}

@article{jiang_language_2019,
 author = {Yiding Jiang and
Shixiang Gu and
Kevin Murphy and
Chelsea Finn},
journal = {Proc. of NeurIPS},
 title = {Language as an Abstraction for Hierarchical Deep Reinforcement Learning},
 year = {2019}
}

@article{ahn2022can,
  title={Do as i can, not as i say: Grounding language in robotic affordances},
  author={Ahn, Michael and Brohan, Anthony and Brown, Noah and Chebotar, Yevgen and Cortes, Omar and David, Byron and Finn, Chelsea and Fu, Chuyuan and Gopalakrishnan, Keerthana and Hausman, Karol and others},
  eprint={2204.01691},
  archivePrefix={arXiv},
  primaryClass={cs.CS},
  year={2022}
}

@article{luria1959directive,
 author = {Luria, Aleksander R},
 journal = {Word},
 number = {2},
 publisher = {Taylor \& Francis},
 title = {The Directive Function of Speech in Development and Dissolution},
 volume = {15},
 year = {1959}
}

@article{roy2022explainability,
  title={Explainability Via Causal Self-Talk},
  author={Roy, Nicholas A and Kim, Junkyung and Rabinowitz, Neil},
  journal={Advances in Neural Information Processing Systems},
  volume={35},
  pages={7655--7670},
  year={2022}
}

@article{hu2023thought,
  title={Thought cloning: Learning to think while acting by imitating human thinking},
  author={Hu, Shengran and Clune, Jeff},
  journal={Proc. of NeurIPS},
  volume={36},
  year={2023}
}

@article{klissarov2023motif,
  title={Motif: Intrinsic motivation from artificial intelligence feedback},
  author={Klissarov, Martin and D'Oro, Pierluca and Sodhani, Shagun and Raileanu, Roberta and Bacon, Pierre-Luc and Vincent, Pascal and Zhang, Amy and Henaff, Mikael},
eprint={2310.00166},
      archivePrefix={arXiv},
      primaryClass={cs.CS},
  year={2023}
}
\bibliographystyle{tmlr}

\appendix
\section{Appendix}
\section{Appendix}
\subsection{Hyperparameters} \label{app:hyperparams}

\begin{enumerate}
\item \textsc{SaySelect}: We use the same default parameters used in InstructRL  \citep{hu2023instructrl}, including setting the regularization strength $\lambda=0.25$. Additionally, each time we invoke \textsf{gen\_rule}, we create an ensemble of 3 rules, which we aggregate over when construct the probability distribution over actions during \textsf{update}.  
    \item \textsc{Maze}: We use the InstructRL objective with a tabular Q-learning agent, but introduce a $\epsilon_{LM}$ parameter that controls whether the regularization strength $\lambda$ is 0 or 1 at each time-step in an episode. Although we did not observe a strong effect on modifying $\epsilon_{LM}$, we did not explore values larger than  $\epsilon_{LM}=0.4$ as that led to increased inference cost and experiment latency. Finally, each time we invoke \textsf{gen\_rule} we create an ensemble of 4 rules, which we aggregate over when construct the probability distribution over actions during \textsf{update}. 
\item \textsc{Builder}: For the listener agent we finetune BART for a maximum of 100 epochs at each iteration of \rllb, employing early stopping on a held-out validation set and using the default arguments provided by the HuggingFace Transformers library \citep{lewis2019bart}. Each time we invoke  \textsf{gen\_rule}, we same 3 rules and select the rule with the highest aggregate likelihood across all tokens. Likewise, for each image description (the speaker's action), we sample 3 possible descriptions under the given rule and select the one with the highest probability. 
\item \textsc{Birds}: For the fine-tuned version of the listener agent,  we finetune StableDiffusion for 1000 steps on a separate finetuning dataset. When generating images, we simply sample one image per description, using 10 inference steps and a guidance scale of 7.5.
\item \textsc{Grasp:} For grasp keypoint selection,we  train an upper confidence bound agent on a neural contextual bandit using image features as context, with learning rate 0.01, with an $(100, 100)$ action space containing integer coordinates of the environment image. During \textsf{update}, we sample 20 coordinates following a given rule from \textsf{gen\_rule}, which we use as the probability distribution over actions, setting $\lambda=0.2$ as the regularization strength. To execute the grasp, we use a grasp model built by training \textsf{Sphinx} \citep{sundaresan2025whats} over 50 waypoint trajectories with learning rate 0.0001 for 100 epochs. 
\end{enumerate}

\subsection{SelectSay Overview}
\begin{figure}[t]
    \centering
    \includegraphics[width=0.4\textwidth]{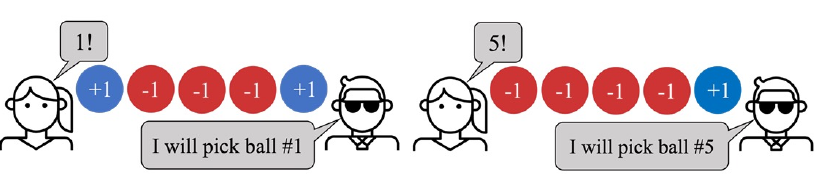}
    \caption{Overview of \textsc{SaySelect} game, reproduced from \citet{hu2023instructrl}.}
    \label{fig:ss-overview}
\end{figure}

\subsection{Maze} \label{app:figs}
\begin{figure}[h]
    \centering
    \includegraphics[width=0.35\textwidth]{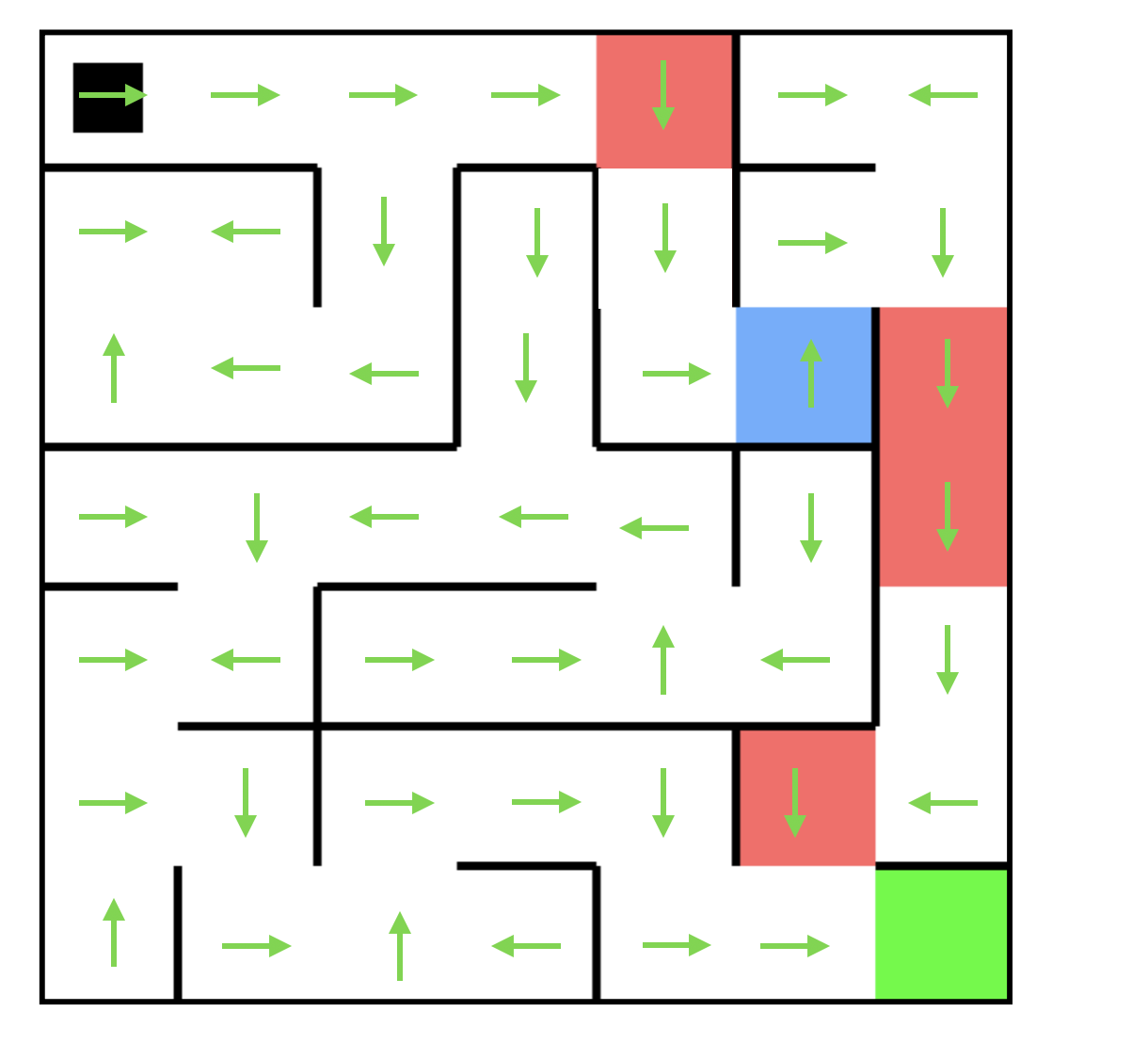}
    \caption{Visual aid provided to participants in the human subject study for the \textsc{Maze} task.}
    \label{fig:maze-visualaid}
\end{figure}

\begin{figure}[h]
    \centering
    \includegraphics[width=0.30\textwidth]{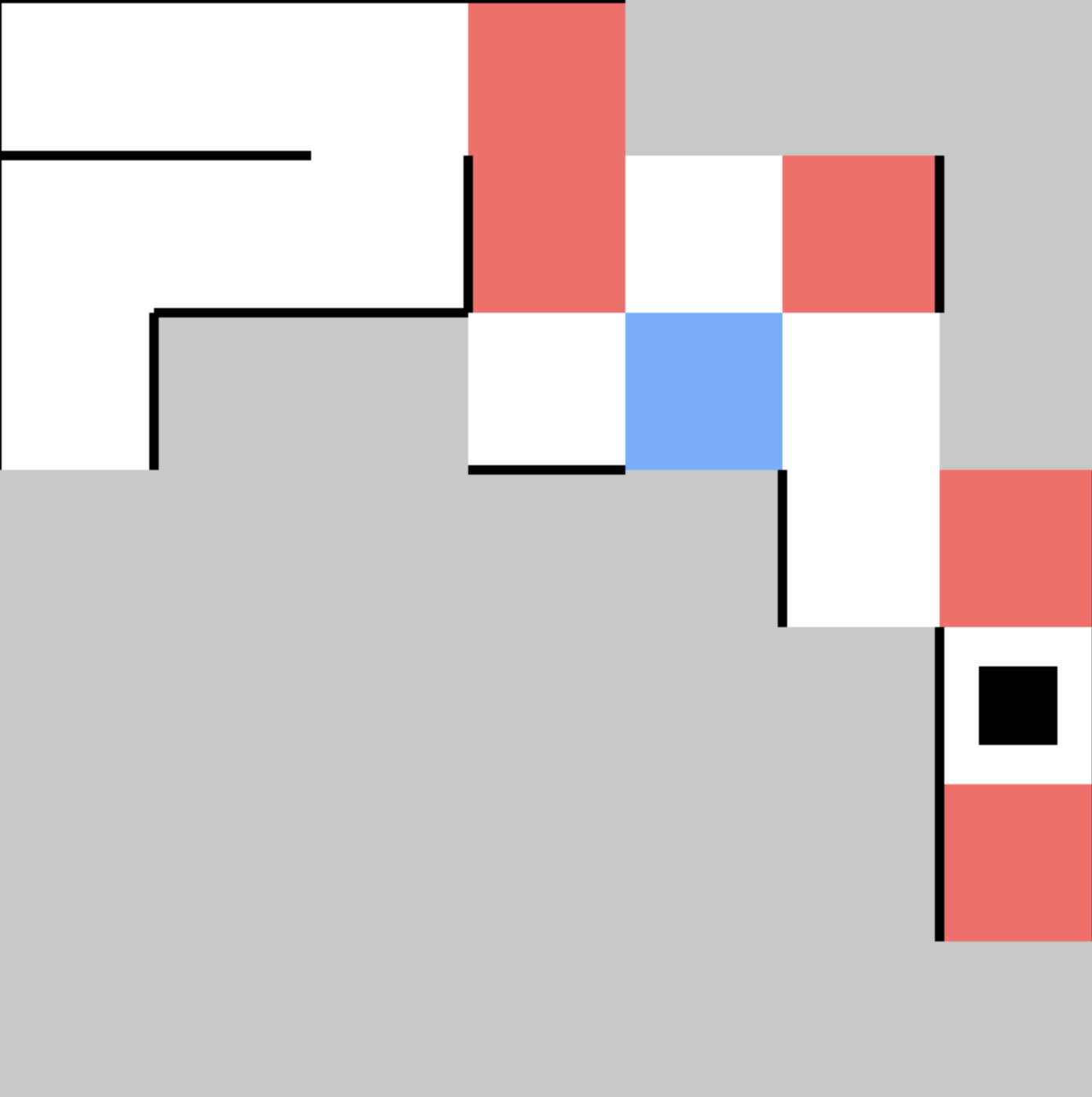}
    \caption{Interface used in the human subject study for the \textsc{Maze} task. Cells not visited by the participants are hidden in gray.}
    \label{fig:maze-interface}
\end{figure}

See Figures \ref{fig:maze-visualaid} and \ref{fig:maze-interface}.

\subsection{Dataset Details for \textsc{Birds}.}\label{app:birds_dataset}
We consider images of the following bird species in the CUB-200-2011 dataset from \citet{cub2011birds} for the \textsc{Birds} task: \texttt{[Indigo Bunting, Cardinal, Yellow Breasted Chat, American Crow, Vermillion Flycatcher, California Gull, Blue Jay, Tropical Kingbird, White Pelican, Horned Puffin]}.

\begin{figure*}[h]
    \centering
    \includegraphics[width=0.95\textwidth]{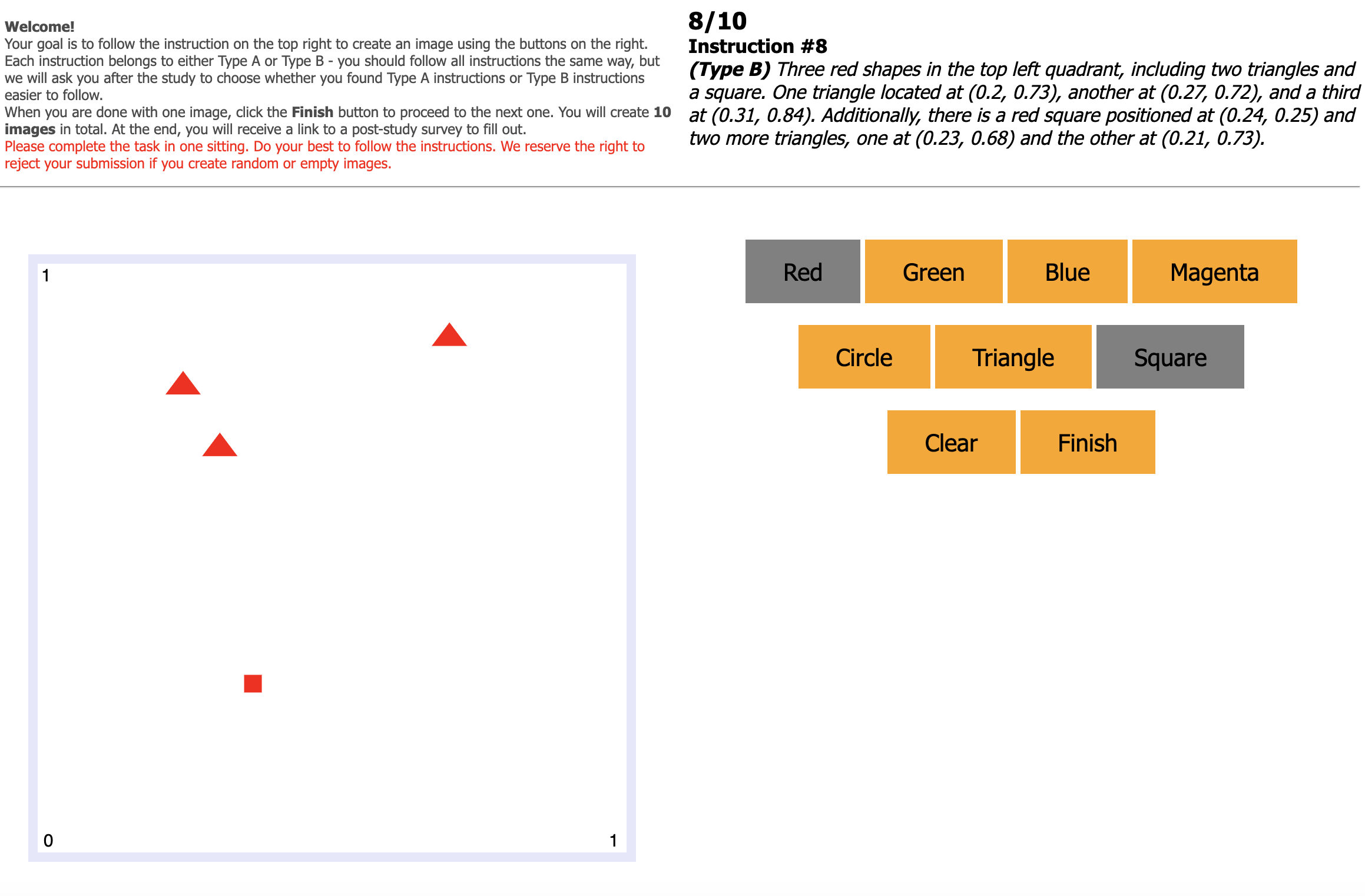}
    \caption{Interface used in the human subject study for the \textsc{Builder} task.}
    \label{fig:builder-interface}
\end{figure*}

\subsection{Grasp}
\label{graspenv}

\begin{figure}[h]
    \centering
    \includegraphics[width=0.45\textwidth]{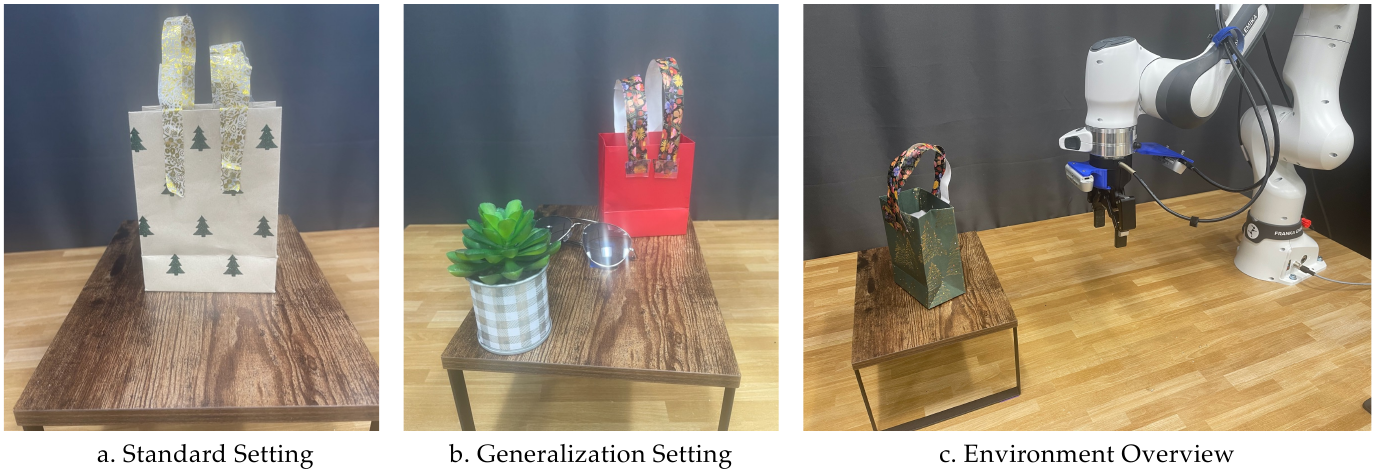}
    \caption{Task environment for \textsc{Grasp}, showing new location and distractors for the Generalization setting.}
    \label{fig:robot-setup}
\end{figure}

\subsection{Builder and Birds Iterations}
\begin{figure}[t]
    \centering
    \includegraphics[width=0.47\textwidth]{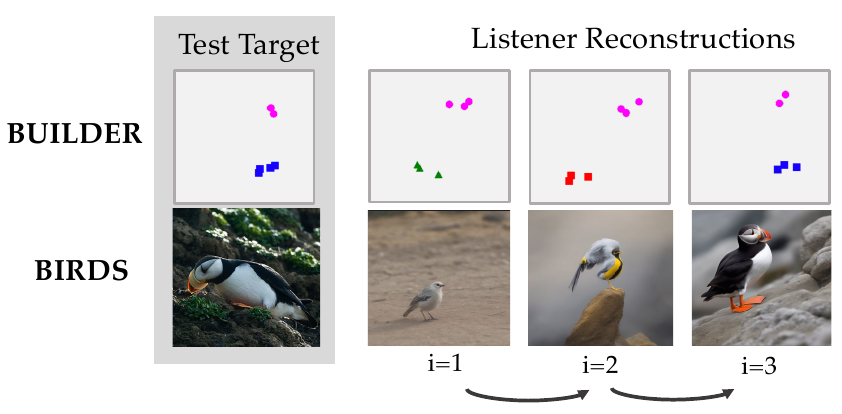}
    \caption{Both the \textsc{builder} and \textsc{birds} tasks consist of speaker and listener agents. At test time, the speaker needs to provide a language description to the listener that helps them recreate the image accurately. For both tasks,  \rllb helps improve listener accuracy over time.}
    \label{fig:builder-bird-overivew}
\end{figure}
\subsection{Full Prompts} \label{app:prompts}
We first provide the full prompts used for \texttt{gen\_rule}. For space constraints, we do not include example \texttt{samples}, but note they follow the format shown in Figure \ref{fig:tasks}.

\begin{enumerate}
\small
    \item \textsc{SaySelect}: You will be given a list of (OBSERVATION, ACTION, REWARD) examples collected from two agents learning to solve a task. Possible ACTIONS an agent can take are: 1, 2, 3, 4, 5, and quit. Each OBSERVATION describes the ordered sequence of actions that AGENT 1 picks, and each ACTION describes the ACTION that AGENT 2 picks based on the given OBSERVATION.
The examples are separated into HIGH REWARD and LOW REWARD examples.+\texttt{[samples]}+Output a language rule that best summarizes the strategy AGENT 2 should follow to receive HIGH REWARD, not LOW REWARD, based on the examples. Start the instruction with the prefix 'I should'. 

\item \textsc{Maze}: You will be given a list of example (OBSERVATION, ACTION) trajectories collected from an AGENT learning to solve a maze. 
Each trajectory receives a REWARD.
Possibles OBSERVATIONS an agent see are: WHITE, RED, BLUE
Possible ACTIONS an agent can take are: NORTH, SOUTH, EAST, WEST. 
The examples are separated into HIGH REWARD and LOW REWARD examples + \texttt{[samples]} + 
Output a language rule that best summarizes the strategy the AGENT should follow when picking a sequence of ACTIONS to solve the maze and receive HIGH REWARD, not LOW REWARD, based on the examples.
Start the instruction with the prefix 'I should'. 

\item \textsc{Builder}: There are two agents. The goal of Agent 1 is to provide instructions to Agent 2 that helps Agent 2 to successfully recreate the image. You will be given a list of (ORIGINAL, AGENT 1 INSTRUCTION, REWARD) values where ORIGINAL is the original description of an image, INSTRUCTION is the instruction provided by Agent 1 to Agent 2, and REWARD is the reward Agent 2 receives when trying to re-create the image (higher is better). 
The examples are separated into HIGH REWARD and LOW REWARD examples.  + \texttt{[samples]}+

Based on the examples above, output a  list of 2 RULES for Agent 1 to follow when generating INSTRUCTION in order to receive HIGH REWARD, instead of LOW REWARD.
Write the rules after the prefix RULES:

\item \textsc{Birds}:  The top row of three images have the following HIGH REWARD descriptions:+\texttt{high reward samples}+The bottom row of three images have the following LOW REWARD descriptions:+\texttt{low reward samples}+Provide a rule I should follow in order to provide image descriptions with HIGH REWARD, not LOW REWARD. Provide the rule after the prefix RULE:

\item \textsc{Grasp}:  The top image shows a grasp keypoint with HIGH REWARD. The bottom image shows a grasp keypoint with LOW REWARD. Based on these images, provide a rule the robot should follow in order to select a grasp keypoint that results in HIGH REWARD, not LOW REWARD. Provide the rule after the prefix RULE: 

\end{enumerate}

We next provide the full prompts used in \texttt{update} for each task.

\begin{enumerate}
\small
    \item \textsc{SaySelect}: $[\mathcal{L}]$+Agent 1 selected \texttt{[observation]}. So I should select
\item \textsc{Maze}: You are an agent solving a maze following a provided RULE. You will be given a list of PREVIOUS ACTIONS and the CURRENT OBSERVATION. Follow the RULE to select your NEXT ACTION (East, West, South, North):

RULE: + $[\mathcal{L}]$+
PREVIOUS ACTIONS: + $[\tau_{1...t-1}]$ +
CURRENT OBSERVATION: + \texttt{[observation]} +

What is the NEXT ACTION you should take? Output one of (East, West, South, North) after the prefix NEXT ACTION:.
\item \textsc{Builder}: You will be given a DESCRIPTION of an image. Your goal is to use this description to provide a short INSTRUCTION to help someone else, who cannot see the image, accurately re-construct it. You will also be given a list of RULES you must follow when providing the instruction.

DESCRIPTION: + \texttt{observation} +

RULES: +$\mathcal{L}$ +

Please provide a short instruction following the prefix INSTRUCTION:
\item \textsc{Birds}: Provide a one-sentence description of this image, using the following RULES:+$\mathcal{L}$

\item \textsc{Grasp}: Provide a keypoint in the image where the robot should grasp the object, following the RULE: $\mathcal{L}$. 
\end{enumerate}

\newpage

\subsection{Examples of Generated Rules for All Environments} \label{app:rules}
See Tables \ref{tab:selectsay_chains}, \ref{tab:maze_chains}, \ref{tab:builder_chains}, \ref{tab:diffusion_chains}, and \ref{tab:grasp_chains}.

\begin{table*}[h]
\footnotesize	
    \centering
    \begin{tabular}{|c|p{40mm}|p{40mm}|p{40mm}|}
    \hline
          Setting & $\mathcal{L}_{1}$ & $\mathcal{L}_{2}$  & $\mathcal{L}_{5}$  \\ \hline
         Standard  & I should follow the strategy of choosing action 4 whenever the observation is 2, 3, 4, or 5.
   &  I should follow the strategy of choosing actions based on the observation provided by AGENT 1. If the observation is 1, I should choose action 1. If the observation is 2, I should choose action 3. If the observation is 3, I should choose action 3. If the observation is 4, I should choose action 4. If the observation is 5, I should choose action 5.&I should follow the strategy of choosing the same action as AGENT 1 for observations 1, 2, 3, 4, and 5.\\ \hline 
         Fixed Speaker &  I should choose action 1 when observation is 1 or 2 or 4 or 5. I should choose action 2 when observation is 3. & I should choose action 1 when observation is 1, 2, 3 or 5. I should choose action 2 when observation is 4.  & I should choose action 1 when observation is 1.
I should choose action 4 when observation is 2.
I should choose action 5 when observation is 3.
I should choose action 2 when observation is 4.
I should choose action 3 when observation is 5.\\ \hline
    \end{tabular}
    \caption{Example $\mathcal{L}$ rules generated for the \textsc{SaySelect} environment, for the Standard setting (both Listener and Speaker agents are RL agents trained from random initialization) and a Fixed Speaker agent.}
    \label{tab:selectsay_chains}
\end{table*}

\begin{table*}[t]
\footnotesize
\centering
    \begin{tabular}{|p{13mm}|p{33mm}|p{33mm}|p{33mm}|p{33mm}|}
    \hline
           Size & $\mathcal{L}_{1}$ & $\mathcal{L}_{2}$ & $\mathcal{L}_{5}$ & $\mathcal{L}_{8}$  \\ \hline
         7x7 & I should start by moving north, then continue moving north until I reach a wall, then turn east and move until I reach a wall, then turn south and move until I reach a wall, then turn west and move until I reach the end of the maze.  & I should prioritize moving North, East, and South over moving West, as the HIGH REWARD EXAMPLES show a higher frequency of these actions. 2. I should avoid moving West unnecessarily, as it is rarely seen in the HIGH REWARD EXAMPLES and often leads to LOW REWARD EXAMPLES & I should take the EAST action when I see WHITE, and then repeat the EAST action until I see RED, then take the SOUTH action, and then repeat the EAST action until I see BLUE, then take the NORTH action, then take the EAST action until I see RED again, then take the SOUTH action, and finally take the WEST action. & I should take the EAST action when I observe WHITE, and keep taking EAST actions until I observe RED, then take the SOUTH action. After taking the SOUTH action, I should take the EAST action again until I observe BLUE, then take the NORTH action. After taking the NORTH action, I should take the EAST action until I observe RED again, then take the SOUTH action. \\ \hline
    \end{tabular}
    \caption{Example $\mathcal{L}$ rules generated for the \textsc{Maze} environment for the 7x7 maze size.}
    \label{tab:maze_chains}
\end{table*}

\begin{table*}[h]
\footnotesize
    \centering
    \begin{tabular}{|c|p{40mm}|p{40mm}|p{40mm}|}
    \hline
          Setting & $\mathcal{L}_{1}$ &  $\mathcal{L}_{2}$ &  $\mathcal{L}_{3}$  \\ \hline
         Re-Initialization &  1. Be specific with location details: Agent 1 should provide detailed location information for each element in the image, such as x and y coordinates. 2. Use descriptive language for elements, such as "red dot" or "green triangle". & 1. Use specific coordinates when instructing Agent 2 to draw shapes. 2. Use descriptive language to specify the color and shape of each element. For example, "a green triangle" instead of "a green thing". & 1. Be specific with location coordinates: provide specific coordinates for the location of each shape, using the format x=0.XX, y=0.YY. 2. Use descriptive shape names: Instead of using generic terms like "dot" or "square," use more descriptive names that indicate the shape's color and size, such as "green triangle" or "red square."\\ \hline
         Continual Training & 1. Be specific and detailed in your instructions. High reward examples have specific coordinates and shapes, while low reward examples have more general descriptions. 2. Use a consistent format for your instructions. High reward examples have a consistent format for listing coordinates and shapes, while low reward examples have a more free-form format. & 1. Provide explicit coordinates for each element in the image, using the format (x, y). 2. Use specific colors when referring to elements in the image, such as "red", "green", or "blue". Avoid using vague terms like "colored" or "shaded". & 1. Use a consistent format for describing shapes, such as always listing the x-coordinate first, followed by the y-coordinate. For example, instead of "one green square at the point x=0.53, y=0.24", use "one green square at (0.53, 0.24)". 2. Avoid using vague terms like "various shades of green". Instead, use specific colors, such as "green" or "blue". Additionally, use specific shapes, such as "square" or "triangle", rather than vague terms like "rectangle". \\ \hline
    \end{tabular}
    \caption{Example $\mathcal{L}$ rules generated for the \textsc{Builder} environment. }
    \label{tab:builder_chains}
\end{table*}

\begin{table*}[h]
\footnotesize
    \centering
    \begin{tabular}{|p{20mm}|p{40mm}|p{40mm}|p{40mm}|}
    \hline
           Reward & $\mathcal{L}_{1}$ & $\mathcal{L}_{2}$ & $\mathcal{L}_{3}$  \\ \hline 
         \textbf{color}  & Describe  the  bird's  color,  species,  and  any  distinctive  markings  or  patterns.  &  Describe  the  bird's  coloration  accurately. & Describe  the  bird's  coloration  accurately.\\ \hline
         \textbf{background}  & Include  details  about  the  bird's  surroundings,  such  as  the  type  of  branch  or  post  it  is  on,  and  any  additional  elements  in  the  background.  & Include  the  bird's  action  (perched,  flying,  standing)  and  its  location  (on  a  branch,  railing,  pole,  etc.) & Describe  the  bird's  action  (flying,  perching,  standing)  and  the  environment  it  is  in  (sky,  tree,  water). \\ \hline
         \textbf{species} & Describe  the  bird's  color,  markings,  and  any  distinctive  features.   & Describe  the  subject's  unique  features,  such  as  coloration,  beak  shape,  or  other  distinguishing  characteristics. & Include  specific  details  about  the  bird's  appearance,  such  as  the  color  of  its  feathers,  beak,  or  eyes,  and  any  distinctive  markings  or  patterns.\\ \hline
    \end{tabular}
    \caption{Example $\mathcal{L}$ rules generated for the \textsc{Birds} environment demonstrate reward-specificity over time. }
    \label{tab:diffusion_chains}
\end{table*}

\begin{table*}[h]
\footnotesize
    \centering
    \begin{tabular}{|p{20mm}|p{40mm}|p{40mm}|p{40mm}|}
    \hline
           Reward & $\mathcal{L}_{10}$ & $\mathcal{L}_{30}$ & $\mathcal{L}_{50}$  \\ \hline 
         \textbf{Shoulder Torque}  &  Select keypoints that are away from sharp edges or corners of the object to avoid potential damage and improve stability during grasping. &  Prioritize grasp keypoints that are closer to the center of the object for more stable and precise pickup. & Prioritize grasping at the middle of the object's surface rather than the top edges or handles to ensure a stable grip.\\ \hline

    \end{tabular}
    \caption{Example $\mathcal{L}$ rules generated for the \textsc{Grasp} environment. }
    \label{tab:grasp_chains}
\end{table*}

\subsection{Effect of Model on Generated Rules} \label{app:rules-qualitative}
We additionally compare the effect of different language models, in particularly those of different sizes, on our results. We did not observe a significantly strong quantitative difference in performance when  \texttt{\textbf{gen\_rule}} is instantiated with models of different sizes (e.g. \textsf{llama-2-70b-chat} vs. \textsf{llama-2-13b-chat}). However, we did notice interesting qualitative differences across samples that are likely due to the additional fine-tuning step using reinforcmenet learning from human feedback (RLHF, see \url{https://llama.meta.com/llama2/} for more information). We describe these differences per environment below. 

\begin{enumerate}
    \item \textsc{SaySelect}: Because of the simplicity of this environment, it is possible for a rule $\mathcal{L}$ to summarize the full optimal human-interpretable policy as a sequence of if-statements (e.g. \textit{``If the current state is '1', I should take action 1.''}) - we observed that smaller language models (e.g. \textsf{llama-2-7b-chat} and \textsf{llama-2-13b-chat}) always did this, while larger models (e.g. \textsf{llama-2-70b-chat} and \textsf{mistral-8x-7b-instruct}) were better able to generalize and use more efficient language, such as \textit{``I should take the same action as the observation''}. The smaller models also often included superfluous language, such as \textit{``I should always take the action that leads to the highest reward''}.   
    \item \textsc{Maze}: The majority of generated rules captured the underlying color semantics of the maze, enabling generalization. However, smaller model sizes (e.g. \textsf{llama-2-7b-chat}) resulted in more superfluous language (e.g. \textit{``I should always prioritize taking actions that lead to the most recent reward, and avoid taking actions that lead to low reward.''}), and the generated rules focused more on the first few actions the agent should take, which may not always generalize. 
    \item \textsc{Builder}: While we do not observe any model-specific differences, there exists variation across samples in the type of formatting and syntax generated rules encourage (e.g. provide coordinates \textit{`` using the format (x, y)''} vs. \textit{`` using the format "x=0.XX, y=0.YY"''}), leading to agents converging to different descriptions. Furthermore, some rules encourage list formats in image descriptions (e.g. \textit{1.Draw a green dot at (0.72, 0.21). 2. Draw a green dot at (0.73, 0.72).}) while other rules encouraged clustering of identical shapes  (e.g.  \textit{Draw two green dots at (0.72, 0.21) and  (0.73, 0.72).})
    \item \textsc{Birds}: Rules demonstrated more reward-specificity (i.e. specific to background, color, or species rewards) when generated with larger VLMs (e.g. \textsf{llava-13b})  versus smaller models (e.g. \textsf{llava-v1.6-vicuna-7b}), with the latter primarily proposing rules that encouraged more detailed descriptions (e.g. \textit{``Avoid using vague or general terms''}).
    \item \textsc{Grasp}: We found that most VLMs other than \textsf{molmo-7B} performed poorly at recognizing keypoints, and instead relied heavily on task information. For example, one rule generated by \textsf{llava-13b} was \textit{``A robot should grasp a keypoint that is visible and not obstructed by other objects.''} We believe the superior performance of the \textsf{molmo} series of models are due to specifically training on the \textsf{PixMo} dataset with 2D-points \citep{allenai2025molmo}. 
\end{enumerate} 

\subsection{Sensitivity Analysis of Rule Generation} \label{app:sensitivity}
We next compare the sensitivity of \rllb to the implementation of \texttt{gen\_rule}, specifically variations in the prompt and temperature for sampling. We present results for \textsc{SaySelect}, as its discrete action space and deterministic dynamics provide the clearest signal for detecting sensitivity effects.

\subsubsection{Prompt Variations}
For \textsc{SaySelect}, we evaluate prompt sensitivity in \texttt{gen\_rule}  by comparing the following variations with the \textbf{Adversarial} baseline described in \ref{sec:selectsay}.:
\begin{itemize}
    \item (\textbf{Original}) You will be given a list of (OBSERVATION, ACTION, REWARD) examples collected from
two agents learning to solve a task. Possible ACTIONS an agent can take are: 1, 2, 3, 4, 5, and quit. Each
OBSERVATION describes the ordered sequence of actions that AGENT 1 picks, and each ACTION describes
the ACTION that AGENT 2 picks based on the given OBSERVATION. The examples are separated into
HIGH REWARD and LOW REWARD examples.+[samples]+Output a language rule that best summarizes
the strategy AGENT 2 should follow to receive HIGH REWARD, not LOW REWARD, based on the
examples. Start the instruction with the prefix `I should’.
\item (\textbf{No Format Instruction}) You will be given a list of (OBSERVATION, ACTION, REWARD) examples collected from
two agents learning to solve a task. Possible ACTIONS an agent can take are: 1, 2, 3, 4, 5, and quit. Each
OBSERVATION describes the ordered sequence of actions that AGENT 1 picks, and each ACTION describes
the ACTION that AGENT 2 picks based on the given OBSERVATION. The examples are separated into
HIGH REWARD and LOW REWARD examples.+[samples]+Output a language rule that best summarizes
the strategy AGENT 2 should follow to receive HIGH REWARD, not LOW REWARD, based on the
examples.
\item (\textbf{Low Context}) You will be given a list of (OBSERVATION, ACTION, REWARD) examples collected from
two agents learning to solve a task. Output a language rule that best summarizes
the strategy AGENT 2 should follow to receive HIGH REWARD, not LOW REWARD, based on the
examples. Start the instruction with the prefix `I should’.
\item (\textbf{Rephrase}) You've been given a list of (OBSERVATION, ACTION, REWARD) triples  from
two agents learning to solve a task. Possible ACTIONS each agent might take are: 1, 2, 3, 4, 5, and quit. Each
OBSERVATION refers to the ordered sequence of actions that AGENT 1 selects, and each ACTION refers to
the ACTION that AGENT 2 selects based on the seen OBSERVATION. The examples are divided into
HIGH REWARD and LOW REWARD examples.+[samples]+Describe the strategy AGENT 2 uses in HIGH REWARD examples that differs from LOW REWARD examples. Start it with  `I should’.
\end{itemize}

Table \ref{tab:sayselect-sensitivity} and \ref{tab:sayselect-sensitivity-intepretability} show that \rllb is mostly robust to variations in prompt and always outperforms the \textbf{Adversarial} baseline, showing that \rllb indeed places higher weight on the content of the contrasting episodes rather than relying on any specific syntax for rule generation.

\subsubsection{Temperature Variation}
For \textsc{SaySelect}, our default temperature for sampling with \texttt{gen\_rule} is 0.5. We compare with temperatures 0.1 (low diversity) and 0.9 (high diversity), and find only slight decrease in reward at the higher temperature of 0.9 (see Table \ref{tab:sayselect-sensitivity}). Importantly, across all temperatures the average reward and human-interpretability scores are stronger than baselines, indicating robustness towards sampling temperature.

\begin{table}[h]
    \centering
    \begin{tabular}{|c|c|c|c|}
     \hline
         Episode & 500 & 1000 & 6000  \\  \hline  
         Reward (\textbf{Original, temp=0.5}) 
         & $0.63 \pm 0.1$ 
         & $0.8 \pm 0.1$ 
         & \cellcolor{green!20}$0.96 \pm 0.03$ \\  \hline
         Reward (\textbf{Original, temp=0.9}) 
         & $0.47 \pm 0.2$ 
         & $0.68 \pm 0.1$ 
         & \cellcolor{green!20}$0.96 \pm 0.05$ \\  \hline
         Reward (\textbf{Original, temp=0.1}) 
         & \cellcolor{green!20}$0.66 \pm 0.1$ 
         & $0.70 \pm 0.09$ 
         & \cellcolor{green!20}$0.96 \pm 0.02$ \\  \hline
         Reward (\textbf{No Format Instruction}) 
         & $0.51 \pm 0.2$  
         & $0.77 \pm 0.05$ 
         & \cellcolor{green!20}$0.96 \pm 0.02$ \\  \hline
         Reward (\textbf{Low Context}) 
         & $0.56 \pm 0.2$ 
         & $0.73 \pm 0.1$  
         & \cellcolor{green!20}$0.96 \pm 0.02$ \\  \hline
         Reward (\textbf{Rephrase}) 
         & $0.64 \pm 0.1$ 
         & \cellcolor{green!20}$0.81 \pm 0.2$ 
         & \cellcolor{green!20}$0.96 \pm 0.02$ \\  \hline
         Reward (\textbf{Adversarial}) 
         & \cellcolor{red!20}$0.21 \pm 0.1$ 
         & \cellcolor{red!20}$0.45 \pm 0.03$ 
         & \cellcolor{red!20}$0.8 \pm 0.05$ \\  \hline
    \end{tabular}
    \caption{Reward across ablations to probe sensitivity of rule generation to prompt variation and temperature. Results are averages across 5 trials.}
    \label{tab:sayselect-sensitivity}
\end{table}

\begin{table}[h]
    \centering
    \begin{tabular}{|c|c|c|c|}
     \hline
         Episode & 500 & 1000 & 6000  \\  \hline  
         Interpretability (\textbf{Original, temp=0.5}) 
         & $0.77\pm 0.03$ 
         & $0.87\pm 0.04$ 
         & \cellcolor{green!20}$0.97\pm 0.02$ \\  \hline
         Interpretability (\textbf{Original, temp=0.9}) 
         & $0.72\pm 0.05$ 
         & $0.80\pm 0.05$ 
         & $0.93\pm 0.03$ \\  \hline
         Interpretability (\textbf{Original, temp=0.1}) 
         & \cellcolor{green!20}$0.87\pm 0.04$ 
         & $0.90 \pm 0.1$ 
         & $0.95\pm 0.02$ \\  \hline
         Interpretability (\textbf{No Format Instruction}) 
         & $0.67 \pm 0.06$  
         & $0.91\pm 0.03$ 
         & $0.94\pm 0.02$ \\  \hline
         Interpretability (\textbf{Low Context}) 
         & $0.77\pm 0.04$ 
         & $0.84\pm 0.05$ 
         & \cellcolor{green!20}$0.97\pm 0.03$ \\  \hline
         Interpretability (\textbf{Rephrase}) 
         & $0.8\pm 0.04$ 
         & \cellcolor{green!20}$0.97\pm 0.04$ 
         & \cellcolor{green!20}$0.97\pm 0.02$ \\  \hline
         Interpretability (\textbf{Adversarial}) 
         & \cellcolor{red!20}$0.55\pm 0.03$ 
         & \cellcolor{red!20}$0.57\pm 0.03$ 
         & \cellcolor{red!20}$0.65\pm 0.06$ \\  \hline
    \end{tabular}
    \caption{Human interpretability across ablations to probe sensitivity of rule generation to prompt variation and temperature. Results are averages across 5 trials.}
    \label{tab:sayselect-sensitivity-intepretability}
\end{table}

\subsection{Non-Contrastive Variant}
A core feature of \rllb is using contrastive episodes (examples of both low and high reward episodes) to \texttt{gen\_rule}. Here we report a non-contrastive ablation for the \textsc{SaySelect} (RL) and \textsc{Birds} (diffusion-model based communication game) environments, where we only use high reward examples in the prompt to \texttt{gen\_rule}. Our results show that contrastive episodes are important for strong performance. Concretely, for \textsc{SaySelect}, the non-contrastive ablation decreases both task reward and human-interpretability of the RL policy,  across different total number of episodes seen (see Table \ref{tab:sayselect-noncontrastive}). Likewise, for the \textsc{Birds} task, the non-contrastive ablation results in lower reward than \rllb, although still leads to a non-trivial improvement over the Baseline and Adversarial methods (see Table \ref{tab:birdsnoncontrastive}). Upon a closer look, we find that without contrasting episodes (i.e. negative episodes that highlight what behaviors are shared between high and low rewarding strategies), \rllb rules often capture spurious features of the tasks, such as \textit{`` Include  details  about  the  bird's  actions.  This  helps  in  creating  a  more  vivid  and  engaging  description.''}. This subsequently results in highly-specific captions such as a bird \textit{`possibly displaying a territorial or mating display`}  or \textit{`looking at the camera`}, neither of which are relevant for any of our reward functions. Overall, these results highlight the importance of using contrasting episodes to improve \rllb rule equality.

\begin{table}[h]
    \centering
    \begin{tabular}{|c|c|c|c|}
     \hline
         Episode & 500 & 1000 & 6000  \\  \hline  
         Reward (\textbf{Original}) 
         & \cellcolor{green!20}$0.63 \pm 0.1$ 
         & \cellcolor{green!20}$0.8 \pm 0.1$ 
         & \cellcolor{green!20}$0.96 \pm 0.03$ \\  \hline
         Reward (\textbf{Non-contrastive}) 
         & $0.53 \pm 0.02$ 
         & $0.7 \pm 0.2$ 
         & $0.91 \pm 0.03$ \\  \hline
         Reward (\textbf{Adversarial}) 
         & \cellcolor{red!20}$0.21 \pm 0.1$ 
         & \cellcolor{red!20}$0.45 \pm 0.03$ 
         & \cellcolor{red!20}$0.8 \pm 0.05$ \\  \hline 
         \\ \hline
         Interpretability (\textbf{Original}) 
         & \cellcolor{green!20}$0.77\pm 0.03$ 
         & \cellcolor{green!20}$0.87\pm 0.04$ 
         & \cellcolor{green!20}$0.97\pm 0.02$ \\  \hline
         Interpretability (\textbf{Non-contrastive}) 
         & $0.75\pm 0.02$ 
         & $0.8\pm 0.06$ 
         & $0.88\pm 0.02$ \\  \hline
         Interpretability (\textbf{Adversarial}) 
         & \cellcolor{red!20}$0.55\pm 0.03$ 
         & \cellcolor{red!20}$0.57\pm 0.03$ 
         & \cellcolor{red!20}$0.65\pm 0.06$ \\  \hline

    \end{tabular}
    \caption{Reward and human-interpretability scores for contrastive and non-contrastive versions of \rllb in \textsc{SaySelect}. Results are averages across 5 trials.}
    \label{tab:sayselect-noncontrastive}
\end{table}

\begin{table}[ht]
\centering

\begin{subtable}{0.3\textwidth}
\centering
\caption{Color}
\begin{tabular}{lc}
\hline
Method & Score \\
\hline
\textbf{Baseline} & $1.69 \pm 0.5$ \\
\textbf{Bottleneck}      & \cellcolor{green!20}$2.93 \pm 0.4$ \\
\textbf{Non-contrastive}      & $2.64 \pm 0.3$ \\
\textbf{Adversarial}     & \cellcolor{red!20}$1.43 \pm 0.5$ \\
\hline
\end{tabular}
\end{subtable}
\hfill
\begin{subtable}{0.3\textwidth}
\centering
\caption{Background}
\begin{tabular}{lc}
\hline
Method & Score \\
\hline
\textbf{Baseline}  & $1.27 \pm 0.5$ \\
\textbf{Bottleneck}     & \cellcolor{green!20}$2.30 \pm 0.3$ \\
\textbf{Non-contrastive}  & $1.78 \pm 0.4$ \\
\textbf{Adversarial}     & \cellcolor{red!20}$1.01 \pm 0.4$ \\
\hline
\end{tabular}
\end{subtable}
\hfill
\begin{subtable}{0.3\textwidth}
\centering
\caption{Species}
\begin{tabular}{lc}
\hline
Method & Score \\
\hline
\textbf{Baseline} & $1.59 \pm 0.5$ \\
\textbf{Bottleneck}   & \cellcolor{green!20}$2.69 \pm 0.5$ \\
\textbf{Non-contrastive} & $2.51 \pm 0.5$ \\

\textbf{Adversarial}     & \cellcolor{red!20}$1.21 \pm 0.4$ \\
\hline
\end{tabular}
\end{subtable}

\caption{Noncontrastive ablation in comparison with main results for \textsc{Birds} task across 3 reward functions. Results are averaged across 5 trials.}
\label{tab:birdsnoncontrastive}
\end{table}

\end{document}